\def\year{2018}\relax
\documentclass[letterpaper]{article} 
\usepackage{aaai18}  
\usepackage{times}  
\usepackage{helvet}  
\usepackage{courier}  
\usepackage{url}  
\usepackage{graphicx}  
\usepackage{subcaption}
\usepackage{amsmath,amssymb,amsfonts,amsthm}
\usepackage{multirow}
\usepackage{algorithm}
\usepackage{algorithmicx}
\usepackage{algpseudocode}
\nocopyright
\newcommand{\bx}{\mathbf{x}}
\newcommand{\by}{\mathbf{y}}
\newcommand{\bz}{\mathbf{z}}
\newcommand{\bw}{\mathbf{w}}
\newcommand{\bbR}{\mathbb{R}}
\newcommand{\cZ}{\mathcal{Z}}
\newcommand{\bdelta}{\boldsymbol{\delta}}
\newcommand{\sign}{\textnormal{sign}}

\begin{document}
%
\title{EAD: Elastic-Net Attacks to Deep Neural Networks via
Adversarial Examples}
\author{Pin-Yu Chen\textsuperscript{1}\thanks{Pin-Yu Chen and Yash Sharma contribute equally to this work. }, Yash Sharma\textsuperscript{2$*$}\thanks{This work was done during the internship of Yash Sharma and Huan Zhang at IBM T. J. Watson Research Center.},  Huan Zhang\textsuperscript{3$\dag$}, Jinfeng Yi\textsuperscript{4}\thanks{Part of the work was done when Jinfeng Yi was at AI Foundations Lab, IBM T. J. Watson Research Center.}, Cho-Jui Hsieh\textsuperscript{3} \\~\\
\textsuperscript{1}AI Foundations Lab, IBM T. J. Watson Research Center, Yorktown Heights, NY 10598, USA\\
\textsuperscript{2}The Cooper Union, New York, NY 10003, USA \\
\textsuperscript{3}University of California, Davis, Davis, CA 95616, USA\\	
\textsuperscript{4}Tencent AI Lab, Bellevue, WA 98004, USA\\	
pin-yu.chen@ibm.com, ysharma1126@gmail.com, ecezhang@ucdavis.edu, jinfengy@us.ibm.com, chohsieh@ucdavis.edu\\
}

\maketitle
\begin{abstract}
Recent studies have highlighted the vulnerability of deep neural networks (DNNs) to adversarial examples - a visually indistinguishable adversarial image can easily be crafted to cause a well-trained model to misclassify. Existing methods for crafting adversarial examples are based on $L_2$ and $L_\infty$ distortion metrics. However,  despite the fact that $L_1$ distortion accounts for the total variation and encourages sparsity in the perturbation,
little has been developed for crafting $L_1$-based adversarial examples.

In this paper, we formulate the process of attacking DNNs via adversarial examples as an elastic-net regularized optimization problem. Our \textbf{e}lastic-net \textbf{a}ttacks to \textbf{D}NNs (EAD) feature $L_1$-oriented adversarial examples and include the state-of-the-art $L_2$ attack as a special case. Experimental results on MNIST, CIFAR10 and ImageNet show that EAD can yield a distinct set of adversarial examples with small $L_1$ distortion and attains similar attack performance to the state-of-the-art methods in different attack scenarios. More importantly, EAD leads to improved attack transferability and complements adversarial training for DNNs, suggesting novel insights on leveraging $L_1$ distortion in adversarial machine learning and security implications of DNNs.
\end{abstract}

\section{Introduction}
\label{sec_intro}
\begin{figure}[t]
		\centering
		\begin{subfigure}[b]{0.245\linewidth}
			\includegraphics[width=\textwidth]{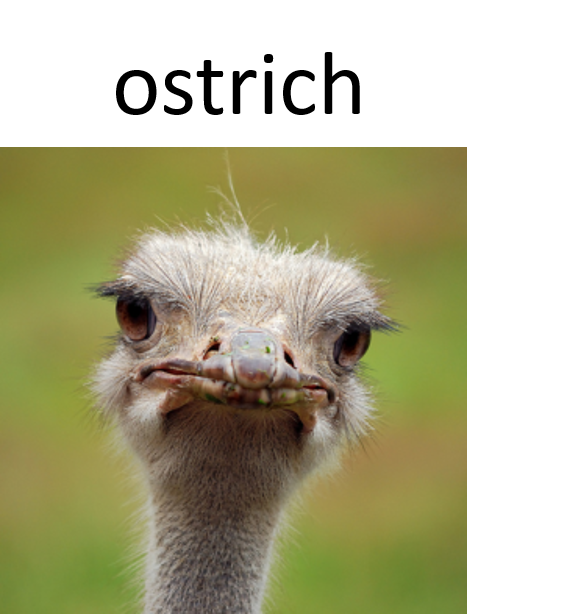}
			\caption{Image}
		\end{subfigure}%
		\centering
		\begin{subfigure}[b]{0.75\linewidth}
			\includegraphics[width=\textwidth]{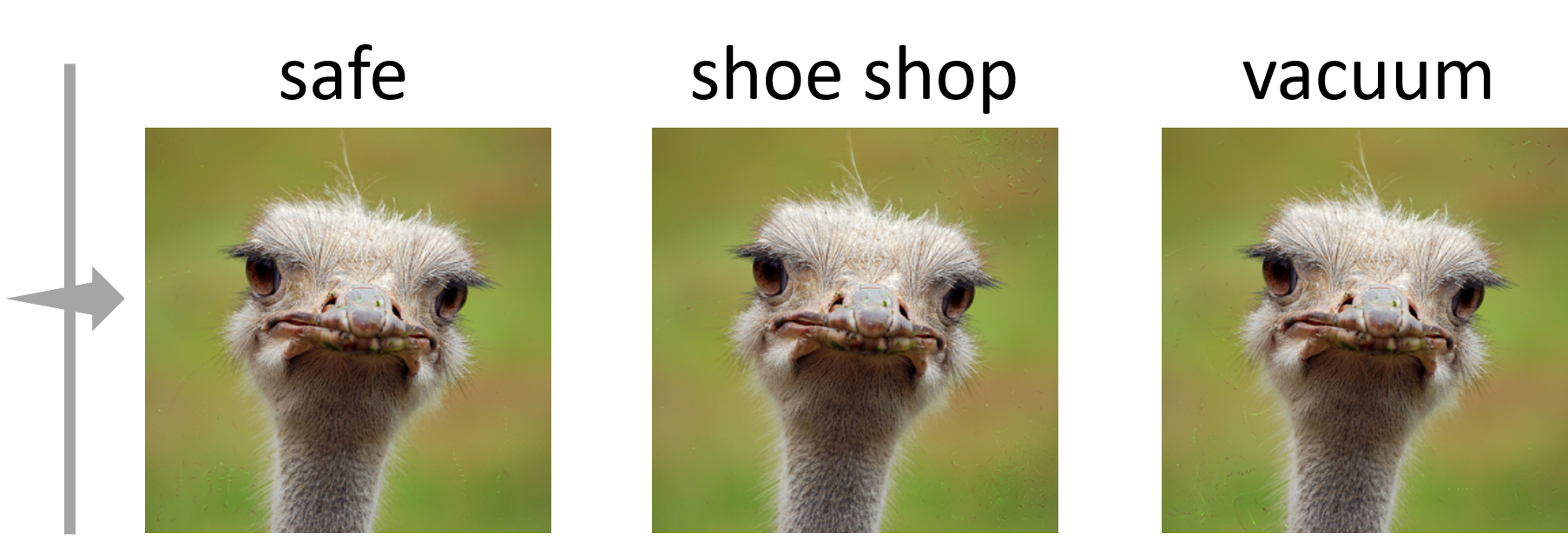}
			\caption{Adversarial examples with target class labels}
		\end{subfigure}
		\caption{Visual illustration of adversarial examples crafted by EAD (Algorithm \ref{algo_EAD}). The original example  is an ostrich image selected from the ImageNet dataset (Figure \ref{Fig_ostrich_demo} (a)). The adversarial examples in Figure \ref{Fig_ostrich_demo} (b) are classified as the target class labels by the Inception-v3 model.}
		\label{Fig_ostrich_demo}
	\end{figure}

Deep neural networks (DNNs) achieve state-of-the-art performance in various tasks in machine learning and artificial intelligence, such as image classification, speech recognition, machine translation and game-playing. Despite their effectiveness, recent studies have illustrated the vulnerability of DNNs to adversarial examples \cite{szegedy2013intriguing,goodfellow2014explaining}. For instance, a carefully designed perturbation to an image can lead a well-trained DNN to misclassify. Even worse, effective adversarial examples can also be made virtually indistinguishable to human perception. For example, Figure \ref{Fig_ostrich_demo} shows three adversarial examples of an ostrich image crafted by our algorithm, which are classified as ``safe'', ``shoe shop" and ``vacuum'' by the Inception-v3 model \cite{szegedy2016rethinking}, a state-of-the-art image classification model.

The lack of robustness exhibited by DNNs to adversarial examples has raised serious concerns for security-critical applications, including traffic sign identification and malware detection, among others. Moreover, moving beyond the digital space, researchers have shown that these adversarial examples are still effective in the physical world at fooling DNNs \cite{kurakin2016adversarial,Evtimov2017robust}. Due to the robustness and security implications, the means of crafting adversarial examples are called \textit{attacks} to DNNs. In particular, \textit{targeted attacks} aim to craft adversarial examples that are misclassified as specific target classes, and \textit{untargeted attacks} aim to craft adversarial examples that are not classified as the original class. \textit{Transfer attacks} aim to craft adversarial examples that are transferable from one DNN model to another. In addition to evaluating the robustness of DNNs, adversarial examples can be used to train a robust model that is resilient to adversarial perturbations, known as \textit{adversarial training} \cite{madry2017towards}. They have also been used in interpreting DNNs \cite{koh2017understanding,dong2017towards}.

Throughout this paper, we use adversarial examples to attack image classifiers based on deep convolutional neural networks.
 The rationale behind crafting effective adversarial examples lies in manipulating the prediction results while ensuring similarity to the original image. Specifically, in the literature the similarity between original and adversarial examples has been measured by different distortion metrics. One commonly used distortion metric is the $L_q$ norm, where $\| \bx \|_q=(\sum_{i=1}^p | \bx_i |^q)^{1/q}$ denotes the $L_q$ norm of a $p$-dimensional vector $\bx=[\bx_1,\ldots,\bx_p]$ for any $q \geq 1$. In particular, when crafting adversarial examples, the $L_\infty$ distortion metric is used to evaluate the maximum variation in pixel value changes \cite{goodfellow2014explaining}, while the $L_2$ distortion metric is used to improve the visual quality \cite{carlini2017towards}. However, despite the fact that the $L_1$ norm is widely used in problems related to image denoising and restoration \cite{fu2006efficient}, as well as sparse recovery \cite{candes2008introduction},  $L_1$-based adversarial examples have not been rigorously explored. In the context of adversarial examples, $L_1$ distortion accounts for the total variation in the perturbation and serves as a popular convex surrogate function of the $L_0$ metric, which measures the number of modified pixels (i.e., sparsity) by the perturbation. To bridge this gap, we propose an attack algorithm based on elastic-net regularization, which we call \textbf{e}lastic-net \textbf{a}ttacks to \textbf{D}NNs (EAD).
Elastic-net regularization is a linear mixture of $L_1$ and $L_2$ penalty functions, and it has been a standard tool for high-dimensional feature selection problems \cite{zou2005regularization}. In the context of attacking DNNs, EAD opens up new research directions since it generalizes the state-of-the-art  attack  proposed in \cite{carlini2017towards} based on $L_2$ distortion, and is able to craft $L_1$-oriented adversarial examples that are more effective and  fundamentally different from existing attack methods.

To explore the utility of $L_1$-based adversarial examples crafted by EAD, we conduct extensive experiments on MNIST, CIFAR10 and ImageNet in different attack scenarios. Compared to the state-of-the-art $L_2$ and $L_\infty$ attacks \cite{kurakin2016adversarial_ICLR,carlini2017towards}, EAD can attain similar attack success rate when breaking undefended and defensively distilled DNNs \cite{papernot2016distillation}. More importantly, we find that $L_1$ attacks attain superior performance over $L_2$ and $L_\infty$ attacks in  transfer attacks and complement adversarial training. 
For the most difficult dataset (MNIST), EAD results in improved attack transferability from an undefended DNN to a defensively distilled DNN, achieving nearly 99\% attack success rate. In addition, joint adversarial training with $L_1$ and $L_2$ based examples can further enhance the resilience of DNNs to adversarial perturbations. These results suggest that EAD yields a distinct, yet more effective, set of adversarial examples. Moreover, evaluating attacks based on $L_1$ distortion provides novel insights on adversarial machine learning and security implications of DNNs, suggesting that $L_1$ may complement $L_2$ and $L_\infty$ based examples toward furthering a thorough adversarial machine learning framework.

\section{Related Work}
\label{sec_related}
Here we summarize related works on attacking and defending DNNs against adversarial examples.

\subsection{Attacks to DNNs}
\noindent
\underline{FGM and I-FGM:} Let $\bx_0$  and $\bx$ denote the original and adversarial examples, respectively, and let $t$ denote the target class to attack.
  Fast gradient methods (FGM) use the gradient $\nabla J$ of the training loss $J$ with respect to $\bx_0$ for crafting adversarial examples \cite{goodfellow2014explaining}. For $L_\infty$ attacks, $\bx$ is crafted by
 \begin{align}
 \label{eqn_FGSM}
 \bx=\bx_0 - \epsilon \cdot \sign (\nabla J (\bx_0,t)),
 \end{align}
 where $\epsilon$ specifies the $L_\infty$ distortion between $\bx$ and $\bx_0$, and  $\sign(\nabla J)$ takes the sign of the gradient. For $L_1$ and $L_2$ attacks, $\bx$ is crafted by
  \begin{align}
  \label{eqn_FGM_L1_L2}
  \bx=\bx_0 - \epsilon \frac{\nabla J (\bx_0,t)}{\|\nabla J (\bx_0,t)\|_q}
  \end{align}
for $q=1,2$, where $\epsilon$ specifies the corresponding  distortion. Iterative fast gradient methods (I-FGM) were proposed in
\cite{kurakin2016adversarial_ICLR}, which iteratively use FGM with a finer distortion, followed by an $\epsilon$-ball clipping. Untargeted attacks using FGM and I-FGM can be implemented in a similar fashion.

\noindent
\underline{C\&W attack:}  Instead of leveraging the training loss, 
Carlini and Wagner designed an $L_2$-regularized loss function based on the logit layer representation in DNNs for crafting adversarial examples \cite{carlini2017towards}. Its formulation turns out to be a special case of our EAD formulation, which will be discussed in the following section. The C\&W attack is considered to be one of the strongest attacks to DNNs, as it can successfully break undefended and defensively distilled DNNs and can attain remarkable attack transferability.

\noindent
\underline{JSMA:}  Papernot et al. proposed a Jacobian-based saliency map algorithm (JSMA) for characterizing the
input-output relation of DNNs \cite{papernot2016limitations}. It can be viewed as a greedy attack algorithm that iteratively modifies the most influential pixel for crafting adversarial examples.

\noindent
\underline{DeepFool:} DeepFool is an untargeted $L_2$ attack algorithm \cite{moosavi2016deepfool} based on the theory of projection to the closest separating hyperplane in classification. It is also used to craft a universal perturbation to mislead DNNs trained on natural images \cite{moosavi2016universal}.

\noindent
\underline{Black-box attacks:} Crafting adversarial examples in the black-box case is plausible if one allows querying of the target DNN. In \cite{papernot2017practical}, JSMA is used to train a substitute model for transfer attacks. In \cite{CPY17_zoo_2}, an effective black-box C\&W attack is made possible using zeroth order optimization (ZOO). In the more stringent attack scenario where querying is prohibited, ensemble methods can be used for transfer attacks \cite{liu2016delving}.

\subsection{Defenses in DNNs}
\noindent
\underline{Defensive distillation:}  Defensive distillation
 \cite{papernot2016distillation} defends against adversarial perturbations by using the distillation technique in \cite{hinton2015distilling} to retrain the same network with class probabilities predicted by the original network. It also introduces the temperature parameter $T$ in the softmax layer to enhance the robustness to adversarial perturbations.

\noindent
\underline{Adversarial training:} Adversarial training can be implemented in a few different ways. A standard approach is augmenting the original training dataset with the label-corrected adversarial examples to retrain the network. Modifying the training loss or the network architecture to increase the robustness of DNNs to adversarial examples has been proposed in \cite{zheng2016improving,madry2017towards,tramer2017ensemble,zantedeschi2017efficient}. 

\noindent
\underline{Detection methods:} Detection methods utilize statistical tests to differentiate adversarial from benign examples \cite{feinman2017detecting,grosse2017statistical,lu2017safetynet,xu2017feature}. However, 10 different detection methods were unable to detect the C\&W attack \cite{carlini2017adversarial}.


\section{EAD: Elastic-Net Attacks to DNNs}
\label{sec_EAD}

\subsection{Preliminaries on Elastic-Net Regularization}
\label{subsec_prelim_EAD}
Elastic-net regularization is a widely used technique in solving
high-dimensional feature selection problems \cite{zou2005regularization}. It can be viewed as a regularizer that linearly combines $L_1$ and $L_2$ penalty functions. In general, elastic-net regularization is used in the following minimization problem:
\begin{align}
\label{eqn_elastic_net}
\textnormal{minimize}_{\bz \in \cZ}~f(\bz)+\lambda_1 \|\bz\|_1 + \lambda_2 \|\bz\|_2^2,
\end{align}
where $\bz$ is a vector of $p$ optimization variables, $\cZ$ indicates the set of feasible solutions, $f(\bz)$ denotes a loss function, $\| \bz \|_q$ denotes the $L_q$ norm of $\bz$,
and $\lambda_1,\lambda_2 \geq 0$ are the $L_1$ and $L_2$ regularization parameters, respectively. The term $\lambda_1 \|\bz\|_1 + \lambda_2 \|\bz\|_2^2$ in (\ref{eqn_elastic_net}) is called the elastic-net regularizer of $\bz$.
 For standard regression problems, 
the loss function $f(\bz)$ is the mean squared error,  the vector $\bz$ represents the weights (coefficients) on the features,
and the set $\cZ=\bbR^p$. In particular, the elastic-net regularization in (\ref{eqn_elastic_net}) degenerates to the LASSO formulation when $\lambda_2=0$, and becomes the ridge regression formulation when $\lambda_1=0$. It is shown in \cite{zou2005regularization}  that elastic-net regularization is able to select a group of highly correlated features, which overcomes the shortcoming of high-dimensional feature selection when solely using the LASSO or ridge regression techniques.

\subsection{EAD Formulation and Generalization}
\label{subsec_formulation_EAD}
Inspired by the C\&W attack \cite{carlini2017towards}, we adopt the same loss function $f$ for crafting adversarial examples. 
Specifically, given an image $\bx_0$ and its correct label denoted by $t_0$, let $\bx$ denote the adversarial example of $\bx_0$ with a target class $t \neq t_0$. The loss function $f(\bx)$ for targeted attacks is defined as 
\begin{align}
\label{eqn_loss_f}
f(\bx,t)=\max \{ \max_{j \neq t} [\textbf{Logit}(\bx)]_j - [\textbf{Logit}(\bx)]_t, - \kappa   \},
\end{align}
where $\textbf{Logit}(\bx)=[[\textbf{Logit}(\bx)]_1,\ldots,[\textbf{Logit}(\bx)]_K] \in \bbR^K$ is the logit layer (the layer prior to the softmax layer) representation of $\bx$ in the considered DNN, $K$ is the number of classes for classification, and $\kappa \geq 0$ is a confidence parameter that guarantees a constant gap between $\max_{j \neq t} [\textbf{Logit}(\bx)]_j$ and 
$[\textbf{Logit}(\bx)]_t$. 

It is worth noting that the term $[\textbf{Logit}(\bx)]_t$ is proportional to the  probability of predicting $\bx$ as label $t$,  since by the softmax classification rule,
\begin{align}
\label{eqn_predict}
\textnormal{Prob}(\textnormal{Label}(\bx)=t)= \frac{\exp([\textbf{Logit}(\bx)]_t)}{\sum_{j=1}^K \exp([\textbf{Logit}(\bx)]_j)}.
\end{align}
Consequently, the loss function in (\ref{eqn_loss_f}) aims to render the label $t$ the most probable class for $\bx$, and the parameter $\kappa$ controls the separation between $t$ and the next most likely prediction among all classes other than $t$. For untargeted attacks, the loss function in (\ref{eqn_loss_f}) can be modified as 
\begin{align}
\label{eqn_loss_f_untargeted}
f(\bx)=\max \{  [\textbf{Logit}(\bx)]_{t_0} - \max_{j \neq t_0} [\textbf{Logit}(\bx)]_j, - \kappa   \}.
\end{align}
In this paper,  we focus on  targeted attacks since 
they are more challenging than untargeted attacks. Our EAD algorithm (Algorithm \ref{algo_EAD}) can directly be applied to untargeted attacks by replacing $f(\bx,t)$ in (\ref{eqn_loss_f}) with $f(\bx)$ in (\ref{eqn_loss_f_untargeted}).

In addition to manipulating the prediction via the loss function in (\ref{eqn_loss_f}),
introducing elastic-net regularization 
further encourages similarity to the original image 
when crafting adversarial examples.  Our formulation of elastic-net attacks to DNNs (EAD) for crafting an adversarial example $(\bx,t)$ with respect to a labeled natural image $(\bx_0,t_0)$ is as follows:
\begin{align}
\label{eqn_EAD_formulation}
 &\textnormal{minimize}_{\bx}~~c \cdot f(\bx,t)+ \beta \|\bx-\bx_0\|_1 +  \|\bx- \bx_0\|_2^2 \nonumber \\
 &\textnormal{subject~to}~~\bx \in [0,1]^p,
\end{align}
where $f(\bx,t)$ is as defined in (\ref{eqn_loss_f}), $c, \beta \geq 0$ are the regularization parameters of the loss function $f$ and the $L_1$ penalty, respectively. The box constraint $\bx \in [0,1]^p$ restricts $\bx$ to a properly scaled image space,  which can be easily satisfied by dividing each pixel value by the maximum attainable value (e.g., 255).
Upon defining the perturbation of $\bx$ relative to $\bx_0$ as $\bdelta=\bx - \bx_0$, the EAD formulation in (\ref{eqn_EAD_formulation}) aims to find an adversarial example $\bx$ that will be classified as the target class $t$ while  minimizing the distortion in $\bdelta$ in terms of the elastic-net loss $\beta \|\bdelta \|_1 +  \|\bdelta \|_2^2$, which is a linear combination of $L_1$ and $L_2$ distortion metrics between $\bx$ and $\bx_0$. 
Notably, the formulation of the C\&W attack \cite{carlini2017towards} becomes a special case of the EAD formulation in (\ref{eqn_EAD_formulation}) when $\beta=0$, which disregards the $L_1$ penalty on $\bdelta$. However, the $L_1$ penalty is an intuitive regularizer for crafting adversarial examples, as $\|\bdelta\|_1=\sum_{i=1}^{p} |\bdelta_i|$ represents the total variation of the perturbation, and is also a widely used surrogate function for promoting sparsity in the perturbation.
As will be evident in the performance evaluation section, including the $L_1$ penalty for the perturbation indeed yields a distinct set of adversarial examples, and it
leads to improved attack transferability and complements adversarial learning.

\subsection{EAD Algorithm}
\label{subsec_algo_EAD}
When solving the EAD formulation in (\ref{eqn_EAD_formulation}) without the $L_1$ penalty (i.e., $\beta=0$), Carlini and Wagner used a change-of-variable (COV) approach via the $\tanh$ transformation on $\bx$ in order to remove the box constraint $\bx \in [0,1]^p$ \cite{carlini2017towards}. When $\beta>0$, we find that the same COV approach is not effective in solving (\ref{eqn_EAD_formulation}), since the corresponding adversarial  examples are insensitive to the changes in $\beta$ (see the performance evaluation section for details).  Since the $L_1$ penalty is a non-differentiable, yet piece-wise linear, function, the failure of the COV approach in solving (\ref{eqn_EAD_formulation}) can be explained by its inefficiency in subgradient-based optimization problems \cite{duchi2009efficient}.

To efficiently solve the EAD formulation in (\ref{eqn_EAD_formulation}) for crafting adversarial examples, we propose to use the iterative shrinkage-thresholding algorithm (ISTA) \cite{beck2009fast}. ISTA can be viewed as a regular first-order optimization algorithm with an additional shrinkage-thresholding step on each iteration. In particular, let $g(\bx)=c \cdot f(\bx)+ \|\bx-\bx_0\|_2^2$ and let  $\nabla g(\bx)$ be the numerical gradient of $g(\bx)$ computed by the DNN. At the $k+1$-th iteration, the adversarial example $\bx^{(k+1)}$ of $\bx_0$ is computed by \begin{align}
\label{eqn_ISTA}
\bx^{(k+1)}=S_{\beta}(\bx^{(k)}-\alpha_k \nabla g(\bx^{(k)}) ),
\end{align}
where $\alpha_k$ denotes the step size at the $k+1$-th iteration, and $S_{\beta}: \bbR^{p} \mapsto \bbR^{p}$ is an element-wise projected shrinkage-thresholding function, which is defined as
\begin{align}
\label{eqn_ISTA_S}
[S_{\beta}(\bz)]_i= \left\{
\begin{array}{ll}
\min \{\bz_i - \beta,1\}, & \text{~if~}\bz_i - {\bx_0}_i > \beta ; \\
{\bx_0}_i, & \text{~if~} |\bz_i - {\bx_0}_i| \leq \beta ; \\
\max \{\bz_i + \beta,0\}, & \text{~if~}\bz_i - {\bx_0}_i < -\beta,
\end{array}
\right.  
\end{align}
for any $i \in \{1,\ldots,p\}$. If $|\bz_i - {\bx_0}_i|> \beta$, it shrinks the element $\bz_i$ by $\beta$ and projects the resulting element to the feasible box constraint between 0 and 1. On the other hand,  if $|\bz_i - {\bx_0}_i| \leq \beta$, it thresholds $\bz_i$ by setting $[S_{\beta}(\bz)]_i = {\bx_0}_i$.
The proof of optimality of using (\ref{eqn_ISTA}) for solving the EAD formulation in (\ref{eqn_EAD_formulation}) is given in the supplementary material\footnote{https://arxiv.org/abs/1709.04114}.
Notably, since $g(\bx)$ is the attack objective function of the C\&W method \cite{carlini2017towards}, the ISTA operation in (\ref{eqn_ISTA}) can be viewed as a robust version of the C\&W method that shrinks a pixel value of the adversarial example if the deviation to the original image is greater than $\beta$, and keeps a pixel value unchanged if the deviation is less than $\beta$.

Our EAD algorithm for crafting adversarial examples is summarized in Algorithm \ref{algo_EAD}. For computational efficiency, a fast ISTA (FISTA) for EAD is implemented, which yields the optimal convergence rate for first-order optimization methods \cite{beck2009fast}. The slack vector $\by^{(k)}$ in Algorithm \ref{algo_EAD} incorporates the momentum in $\bx^{(k)}$ for acceleration. 
In the experiments, we set the initial learning rate $\alpha_0=0.01$ with a square-root decay factor in $k$. During the EAD iterations, the iterate $\bx^{(k)}$ is considered as a successful adversarial example of $\bx_0$ if the model predicts its most likely class to be the target class $t$. The final adversarial example $\bx$ is selected from all successful examples based on distortion metrics. In this paper we consider two decision rules for selecting $\bx$: the least elastic-net (EN) and $L_1$ distortions  relative to $\bx_0$. The influence of $\beta$, $\kappa$ and the decision rules on EAD will be investigated in the following section.

\begin{algorithm}[t]
	\caption{Elastic-Net Attacks to DNNs (EAD)}
	\label{algo_EAD}
	\begin{algorithmic}
		\State \textbf{Input:} original labeled image $(\bx_0,t_0)$, target attack class $t$, attack transferability parameter $\kappa$, $L_1$ regularization parameter $\beta$, step size $\alpha_k$, \# of iterations $I$	
		\State \textbf{Output:} adversarial example $\bx$
		\State Initialization: 	$\bx^{(0)}=\by^{(0)}=\bx_0$
		\For{$k=0$ to $I-1$} 
	\State $\bx^{(k+1)}=S_{\beta}(\by^{(k)}-\alpha_k \nabla g(\by^{(k)}))$ 
	\State $\by^{(k+1)}=\bx^{(k+1)}+\frac{k}{k+3} (\bx^{(k+1)} - \bx^{(k)})$
		\EndFor
	\State Decision rule: determine $\bx$ from successful examples in $\{\bx^{(k)}\}_{k=1}^I$ (EN rule or $L_1$ rule).
	\end{algorithmic} 
\end{algorithm} 

\section{Performance Evaluation}
\label{sec_performance}
In this section, we compare the proposed EAD with the state-of-the-art attacks to DNNs on three image classification datasets -  MNIST, CIFAR10 and ImageNet.
We would like to show that (i) EAD can attain attack performance similar to the C\&W attack in breaking undefended and defensively distilled DNNs, since  the C\&W attack is a special case of EAD when $\beta=0$; (ii)  Comparing to existing 
$L_1$-based FGM and I-FGM methods, the adversarial examples using EAD can lead to significantly lower $L_1$ distortion and better attack success rate; (iii) The $L_1$-based adversarial examples crafted by EAD can achieve improved  attack transferability and complement adversarial training.

\subsection{Comparative Methods}
We compare EAD with the following targeted attacks, which are the most effective methods for crafting adversarial examples in different distortion metrics.

\noindent
\textbf{C\&W attack:} The state-of-the-art $L_2$ targeted attack proposed by Carlini and Wagner \cite{carlini2017towards}, which is a special case of EAD when $\beta=0$. 

\noindent
\textbf{FGM:} The fast gradient method proposed in \cite{goodfellow2014explaining}. The FGM attacks using different distortion metrics are denoted by FGM-$L_1$, FGM-$L_2$ and FGM-$L_\infty$.

\noindent
\textbf{I-FGM:} The iterative fast gradient method proposed in \cite{kurakin2016adversarial_ICLR}.  The I-FGM attacks using different distortion metrics are denoted by I-FGM-$L_1$, I-FGM-$L_2$ and I-FGM-$L_\infty$.

\begin{table*}[t]
	\centering
	\caption{Comparison of the change-of-variable (COV) approach and EAD (Algorithm \ref{algo_EAD}) for solving the elastic-net formulation in (\ref{eqn_EAD_formulation}) on MNIST.  ASR means attack success rate (\%). Although these two methods attain similar attack success rates, COV is not effective in crafting $L_1$-based adversarial examples. Increasing $\beta$ leads to less $L_1$-distorted adversarial examples for EAD, whereas the distortion of COV is insensitive to changes in $\beta$.}
	\label{table_COV_EAD}
\begin{tabular}{ll|llll|llll|llll}
	\hline
	&           & \multicolumn{4}{c|}{Best case}   & \multicolumn{4}{c|}{Average case} & \multicolumn{4}{c}{Worst case}    \\ \hline
	\begin{tabular}[c]{@{}l@{}}Optimization \\ method\end{tabular}           & $\beta$   & ASR & $L_1$ & $L_2$ & $L_\infty$ & ASR  & $L_1$ & $L_2$ & $L_\infty$ & ASR  & $L_1$ & $L_2$ & $L_\infty$ \\ \hline
	\multirow{5}{*}{COV}                                                     & $0$       & 100 & 13.93 & 1.377 & 0.379      & 100  & 22.46 & 1.972 & 0.514      & 99.9 & 32.3  & 2.639 & 0.663      \\
	& $10^{-5}$ & 100 & 13.92 & 1.377 & 0.379      & 100  & 22.66 & 1.98  & 0.508      & 99.5 & 32.33 & 2.64  & 0.663      \\
	& $10^{-4}$ & 100 & 13.91 & 1.377 & 0.379      & 100  & 23.11 & 2.013 & 0.517      & 100  & 32.32 & 2.639 & 0.664      \\
	& $10^{-3}$ & 100 & 13.8  & 1.377 & 0.381      & 100  & 22.42 & 1.977 & 0.512      & 99.9 & 32.2  & 2.639 & 0.664      \\
	& $10^{-2}$ & 100 & 12.98 & 1.38  & 0.389      & 100  & 22.27 & 2.026 & 0.53       & 99.5 & 31.41 & 2.643 & 0.673      \\ \hline
	\multirow{5}{*}{\begin{tabular}[c]{@{}l@{}}EAD\\ (EN rule)\end{tabular}} & $0$       & 100 & 14.04 & 1.369 & 0.376      & 100  & 22.63 & 1.953 & 0.512      & 99.8 & 31.43 & 2.51  & 0.644      \\
	& $10^{-5}$ & 100 & 13.66 & 1.369 & 0.378      & 100  & 22.6  & 1.98  & 0.515      & 99.9 & 30.79 & 2.507 & 0.648      \\
	& $10^{-4}$ & 100 & 12.79 & 1.372 & 0.388      & 100  & 20.98 & 1.951 & 0.521      & 100  & 29.21 & 2.514 & 0.667      \\
	& $10^{-3}$ & 100 & 9.808 & 1.427 & 0.452      & 100  & 17.4  & 2.001 & 0.594      & 100  & 25.52 & 2.582 & 0.748      \\
	& $10^{-2}$ & 100 & 7.271 & 1.718 & 0.674      & 100  & 13.56 & 2.395 & 0.852      & 100  & 20.77 & 3.021 & 0.976      \\ \hline
\end{tabular}
\end{table*}

\subsection{Experiment Setup and Parameter Setting}
Our experiment setup is based on Carlini and Wagner's framework\footnote{\url{https://github.com/carlini/nn_robust_attacks}}. 
For both the EAD and C\&W attacks, we use the default setting\footnotemark[1], which
implements 9 binary search steps on the regularization parameter $c$ (starting from 0.001)
and runs $I=1000$ iterations for each step with the initial learning rate $\alpha_0=0.01$. For finding successful adversarial examples, we use the reference optimizer\footnotemark[1] (ADAM) for the C\&W attack and implement the projected FISTA (Algorithm \ref{algo_EAD}) with the square-root decaying learning rate for EAD. 
Similar to the C\&W attack, the final adversarial example of EAD is selected by the least distorted example among all the successful examples. The sensitivity analysis of the $L_1$ parameter  $\beta$ and the effect of the decision rule on EAD will be investigated in the forthcoming paragraph.
Unless specified, we set the attack transferability parameter $\kappa = 0$ for both attacks.

We implemented FGM and I-FGM using the CleverHans package\footnote{\url{https://github.com/tensorflow/cleverhans}}.  The best distortion parameter $\epsilon$ is determined by a fine-grained grid search - for each image, the smallest $\epsilon$ in the grid leading to a successful attack is reported.  For I-FGM, we perform 10 FGM iterations (the default value) with $\epsilon$-ball clipping. The distortion parameter $\epsilon^\prime$ in each FGM iteration is set to be $\epsilon/10$, which has been shown to be an effective attack setting in \cite{tramer2017ensemble}.
The range of the grid and the resolution of these two methods are specified in the supplementary material\footnotemark[1].

The image classifiers for MNIST and CIFAR10 are trained based on the DNN models provided by Carlini and Wagner\footnotemark[1]. The image classifier for ImageNet is the Inception-v3 model \cite{szegedy2016rethinking}.
For MNIST and CIFAR10, 1000 correctly classified images are randomly selected from the test sets to attack an incorrect class label.
For ImageNet, 100 correctly classified images and 9 incorrect classes are randomly selected to attack.
All experiments are conducted on a machine with an Intel E5-2690 v3 CPU, 40 GB RAM and a single NVIDIA K80 GPU. Our EAD code is publicly available for download\footnote{\url{ https://github.com/ysharma1126/EAD-Attack}}.

\begin{figure*}[t]
	\centering
	\begin{subfigure}[b]{0.25\linewidth}
		\includegraphics[width=\textwidth]{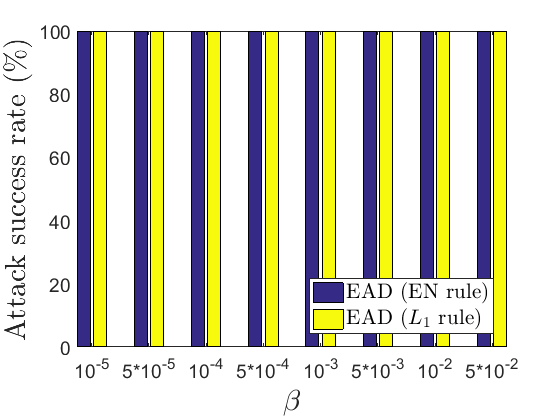}
	\end{subfigure}%
	\centering
	\begin{subfigure}[b]{0.25\linewidth}
		\includegraphics[width=\textwidth]{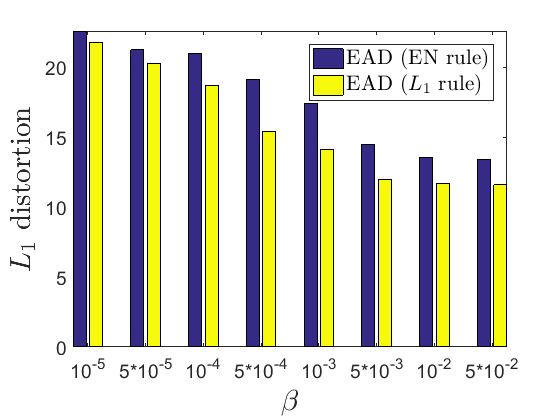}
	\end{subfigure}%
	\centering
	\begin{subfigure}[b]{0.25\linewidth}
		\includegraphics[width=\textwidth]{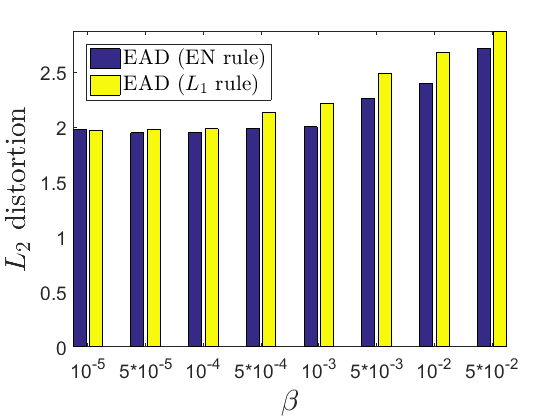}
	\end{subfigure}%
	\centering
	\begin{subfigure}[b]{0.25\linewidth}
		\includegraphics[width=\textwidth]{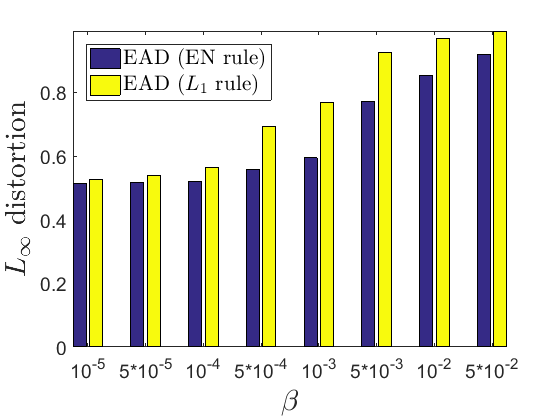}
	\end{subfigure}
	\caption{Comparison of EN and $L_1$ decision rules in EAD on MNIST with varying $L_1$ regularization parameter $\beta$ (average case). Comparing to the EN rule, for the same $\beta$
		 the $L_1$ rule attains less $L_1$ distortion but may incur more  $L_2$ and $L_\infty$ distortions.  }
	\label{Fig_rule_mnist}
\end{figure*}

\subsection{Evaluation Metrics}
Following the attack evaluation criterion in \cite{carlini2017towards}, we report the attack success rate and distortion of the adversarial examples from each method. The attack success rate (ASR) is defined as the percentage of adversarial examples that are classified as the target class (which is different from the original class). The average $L_1$, $L_2$ and $L_\infty$ distortion metrics of successful  adversarial examples are also reported. In particular, the ASR and distortion of the following attack settings are considered:

\noindent
\textbf{Best case:} The least difficult attack among targeted attacks to all incorrect class labels in terms of distortion.

\noindent
\textbf{Average case:} The targeted attack to a randomly selected incorrect class label.

\noindent
\textbf{Worst case:} The most difficult attack among targeted attacks to all incorrect class labels in terms of distortion.

\subsection{Sensitivity Analysis and Decision Rule for EAD}
We verify the necessity of using Algorithm \ref{algo_EAD} for solving the elastic-net regularized attack formulation in (\ref{eqn_EAD_formulation}) by comparing it to a naive change-of-variable (COV) approach. In \cite{carlini2017towards}, Carlini and Wagner remove the box constraint $\bx \in [0,1]^p$ by replacing $\bx$ with $\frac{\mathbf{1}+\tanh{\bw}}{2}$, where $\bw \in \bbR^p$ and $\mathbf{1} \in \bbR^p$ is a vector of ones.  The default ADAM optimizer \cite{kingma2014adam} is then used to solve $\bw$ and obtain $\bx$. We apply this COV approach to (\ref{eqn_EAD_formulation}) and compare with EAD on MNIST with different orders of the $L_1$ regularization parameter $\beta$ in Table \ref{table_COV_EAD}. Although COV and EAD attain similar attack success rates, it is observed that COV is not effective in crafting $L_1$-based adversarial examples. Increasing  $\beta$ leads to less $L_1$-distorted adversarial examples for EAD, whereas the distortion ($L_1$, $L_2$ and $L_\infty$) of COV is insensitive to changes in $\beta$. Similar insensitivity of COV on $\beta$ is observed when one uses other optimizers such as  AdaGrad, RMSProp or built-in SGD in TensorFlow.
We also note that the COV approach prohibits the use of ISTA due to the subsequent $\tanh$ term in the $L_1$ penalty.
The insensitivity of COV suggests that it is inadequate for elastic-net optimization, which can be explained by its inefficiency in subgradient-based optimization problems \cite{duchi2009efficient}. 
For EAD, we also find an interesting trade-off between $L_1$ and the  other two distortion metrics - adversarial examples with smaller $L_1$ distortion tend to have larger $L_2$ and $L_\infty$ distortions. This trade-off can be explained by the fact that increasing $\beta$ further encourages sparsity in the perturbation, and hence results in increased $L_2$ and $L_\infty$ distortion.
Similar results are observed on CIFAR10 (see supplementary material\footnotemark[1]).

\begin{table*}[t]
	\centering
	\caption{Comparison of different attacks on MNIST, CIFAR10 and ImageNet  (average case). ASR means attack success rate (\%).	The distortion metrics are averaged over successful examples.	 EAD, the C\&W attack, and I-FGM-$L_\infty$ attain the least $L_1$, $L_2$, and $L_\infty$ distorted adversarial examples, respectively. The complete attack results are given in the supplementary material\footnotemark[1].}
	\label{table_attack}
	\begin{tabular}{l|llll|llll|llll}
		\hline
		& \multicolumn{4}{c|}{MNIST}        & \multicolumn{4}{c|}{CIFAR10}       & \multicolumn{4}{c|}{ImageNet}    \\ \hline
		Attack method    & ASR  & $L_1$ & $L_2$ & $L_\infty$ & ASR  & $L_1$  & $L_2$ & $L_\infty$ & ASR & $L_1$ & $L_2$ & $L_\infty$ \\ \hline
		C\&W ($L_2$)     & \textbf{100}  & 22.46 & \textbf{1.972} & 0.514      & \textbf{100}  & 13.62  & \textbf{0.392} & 0.044      & \textbf{100} & 232.2 & \textbf{0.705} & 0.03       \\
		FGM-$L_1$        & 39   & 53.5  & 4.186 & 0.782      & 48.8 & 51.97  & 1.48  & 0.152      & 1   & 61    & 0.187 & 0.007      \\
		FGM-$L_2$        & 34.6 & 39.15 & 3.284 & 0.747      & 42.8 & 39.5   & 1.157 & 0.136      & 1   & 2338  & 6.823 & 0.25       \\
		FGM-$L_\infty$   & 42.5 & 127.2 & 6.09  & 0.296      & 52.3 & 127.81 & 2.373 & 0.047      & 3   & 3655  & 7.102 & 0.014      \\
		I-FGM-$L_1$      & \textbf{100}  & 32.94 & 2.606 & 0.591      & \textbf{100}  & 17.53  & 0.502 & 0.055      & 77  & 526.4 & 1.609 & 0.054      \\
		I-FGM-$L_2$      & \textbf{100}  & 30.32 & 2.41  & 0.561      & \textbf{100}  & 17.12  & 0.489 & 0.054      & \textbf{100} & 774.1 & 2.358 & 0.086      \\
		I-FGM-$L_\infty$ & \textbf{100}  & 71.39 & 3.472 & \textbf{0.227}      & \textbf{100}  & 33.3   & 0.68  & \textbf{0.018}      & \textbf{100} & 864.2 & 2.079 & \textbf{0.01}       \\
		EAD (EN rule)    & \textbf{100}  & \textbf{17.4}  & 2.001 & 0.594      & \textbf{100}  & \textbf{8.18}   & 0.502 & 0.097      & \textbf{100} & \textbf{69.47} & 1.563 & 0.238      \\
		EAD ($L_1$ rule) & \textbf{100}  & \textbf{14.11} & 2.211 & 0.768      & \textbf{100}  & \textbf{6.066}  & 0.613 & 0.17       & \textbf{100} & \textbf{40.9}  & 1.598 & 0.293      \\ \hline
	\end{tabular}
\end{table*}

In Table \ref{table_COV_EAD}, during the attack optimization process the final adversarial example is selected based on the elastic-net loss of all successful adversarial examples in $\{\bx^{(k)}\}_{k=1}^I$, which we call the \textit{elastic-net (EN) decision rule}. Alternatively, we can select the final adversarial example with the least $L_1$ distortion, which we call the \textit{$L_1$ decision rule}. Figure \ref{Fig_rule_mnist} compares the ASR and average-case distortion of these two decision rules with different $\beta$ on MNIST. Both decision rules yield 100\% ASR for a wide range of $\beta$ values.
 For the same $\beta$, the $L_1$ rule gives adversarial examples with less $L_1$ distortion than those given by the EN rule at the price of larger $L_2$ and $L_\infty$ distortions. Similar trends are observed on CIFAR10 (see supplementary material\footnotemark[1]). The complete results of these two rules on MNIST and CIFAR10 are given in the supplementary material\footnotemark[1].
In the following experiments, we will report the results of EAD with these two decision rules and set $\beta=10^{-3}$, since on MNIST and CIFAR10 this $\beta$ value significantly reduces the $L_1$ distortion while having comparable $L_2$ and $L_\infty$ distortions to the case of $\beta=0$ (i.e., without $L_1$ regularization). 

\subsection{Attack Success Rate and Distortion on MNIST, CIFAR10 and ImageNet}
We compare EAD with the comparative methods in terms of attack success rate and different distortion metrics on attacking the considered DNNs trained on MNIST, CIFAR10 and ImageNet. 
Table \ref{table_attack} summarizes their average-case performance. It is observed that FGM methods fail to yield successful adversarial examples (i.e., low ASR), and the corresponding distortion metrics are significantly larger than other methods. On the other hand, the C\&W attack, I-FGM and EAD all lead to 100\% attack success rate. Furthermore,  EAD, the C\&W method, and I-FGM-$L_\infty$ attain the least $L_1$, $L_2$, and $L_\infty$ distorted adversarial examples, respectively. We note that EAD significantly outperforms the existing $L_1$-based method (I-FGM-$L_1$). Compared to I-FGM-$L_1$, EAD with the EN decision rule reduces the $L_1$ distortion by roughly 47\% on MNIST, 53\% on CIFAR10 and 87\% on ImageNet. We also observe that EAD with the $L_1$ decision rule can further reduce the $L_1$ distortion but at the price of noticeable increase in the $L_2$ and $L_\infty$ distortion metrics.

Notably, despite having large $L_2$ and $L_\infty$ distortion metrics,  the adversarial examples crafted by  EAD with the $L_1$ rule can still attain 100\% ASRs in all datasets, which implies  the  $L_2$ and $L_\infty$ distortion metrics are insufficient for evaluating the robustness of neural networks. Moreover,
 the attack results in Table \ref{table_attack} suggest that EAD can yield a set of distinct adversarial examples that are fundamentally different from $L_2$ or $L_\infty$ based examples. Similar to the C\&W method and I-FGM, the adversarial examples from EAD are also visually indistinguishable (see supplementary material\footnotemark[1]).

\begin{figure}[t]
	\centering
	\begin{subfigure}[b]{0.5\linewidth}
		\includegraphics[width=\textwidth]{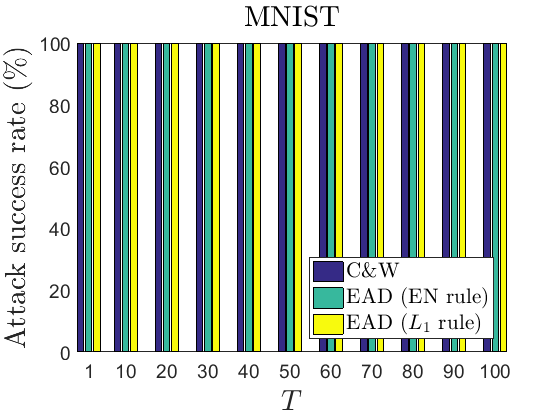}
	\end{subfigure}%
	\centering
	\begin{subfigure}[b]{0.5\linewidth}
		\includegraphics[width=\textwidth]{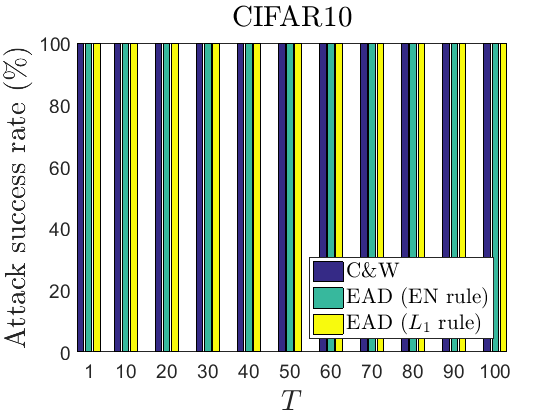}
	\end{subfigure}%
	\caption{Attack success rate (average case) of the C\&W method and EAD on MNIST and CIFAR10 with respect to varying temperature parameter $T$ for defensive distillation. Both methods can successfully break defensive distillation. }
	\label{Fig_DD_ASR}
\end{figure}

\subsection{Breaking Defensive Distillation}
In addition to breaking undefended DNNs via adversarial examples, here we show that EAD can also break defensively distilled DNNs. Defensive distillation \cite{papernot2016distillation} is a standard defense technique that retrains the network with class label probabilities predicted by the original network, soft labels, and introduces the temperature parameter $T$ in the softmax layer to enhance its robustness to adversarial perturbations. Similar to the state-of-the-art attack (the C\&W method),
Figure \ref{Fig_DD_ASR} shows that EAD can attain 100\% attack success rate for different values of $T$ on MNIST and CIFAR10. Moreover, since the C\&W attack formulation is a special case of the EAD formulation in (\ref{eqn_EAD_formulation}) when $\beta=0$, 
 successfully breaking defensive distillation using EAD suggests new ways of crafting effective adversarial examples by varying the $L_1$ regularization parameter $\beta$.
 The complete attack results are given in the supplementary material\footnotemark[1].

\subsection{Improved Attack Transferability}
It has been shown in \cite{carlini2017towards} that the C\&W attack can be  made highly transferable from an undefended network to a defensively distilled network by tuning the confidence parameter $\kappa$ in (\ref{eqn_loss_f}). Following \cite{carlini2017towards}, we adopt the same experiment setting for attack transferability on MNIST, as MNIST is the most difficult dataset to attack in terms of the average distortion per image pixel from Table \ref{table_attack}. 

\begin{figure}[!t]
	\centering
	\includegraphics[width=2.9in]{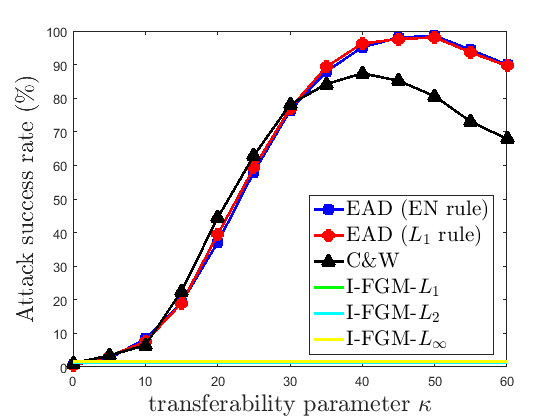}
	\caption{Attack transferability (average case) from the undefended network to the defensively distilled network on MNIST by varying $\kappa$. EAD can attain nearly 99\% attack success rate (ASR) when $\kappa=50$, whereas the top ASR of the C\&W attack is nearly 88\% when $\kappa=40$. }
	\label{Fig_transferability}
\end{figure}

Fixing $\kappa$, adversarial examples generated from the original (undefended) network are used to attack the defensively distilled network with the temperature parameter $T=100$ \cite{papernot2016distillation}. The attack success rate (ASR) of EAD, the C\&W method and I-FGM are shown in Figure \ref{Fig_transferability}. When $\kappa=0$, all methods attain low ASR and hence do not produce transferable adversarial examples. The ASR of EAD and the C\&W method improves when we set $\kappa>0$, whereas I-FGM's ASR remains low (less than 2\%) since the attack does not have such a parameter for transferability. 

Notably, EAD can attain nearly 99\% ASR when $\kappa=50$, whereas the top ASR of the C\&W method is nearly 88\% when $\kappa=40$. This implies improved attack transferability when using the adversarial examples crafted by EAD, which can be explained by the fact that the ISTA operation in (\ref{eqn_ISTA}) is a robust version of the C\&W attack via shrinking and thresholding.
We also find that setting $\kappa$ too large may mitigate the ASR of transfer attacks for both EAD and the C\&W method, as the optimizer may fail to find an adversarial example that minimizes the loss function $f$ in (\ref{eqn_loss_f}) for large $\kappa$. The complete attack transferability results are given in the supplementary material\footnotemark[1].

\begin{table}[t]
	\centering
	\caption{Adversarial training using the C\&W attack and EAD ($L_1$ rule)  on MNIST. ASR means attack success rate. Incorporating $L_1$ examples  complements adversarial training and enhances attack difficulty in terms of distortion.
		 The complete results are given in the supplementary material\footnotemark[1].}
	\label{table_adv_training}
	\begin{tabular}{ll|llll}
		\hline
		\multirow{2}{*}{\begin{tabular}[c]{@{}l@{}}Attack\\ method\end{tabular}}     & \multirow{2}{*}{\begin{tabular}[c]{@{}l@{}}Adversarial\\ training\end{tabular}} & \multicolumn{4}{c}{Average case} \\
		&                                                                                 & ASR & $L_1$ & $L_2$  & $L_\infty$ \\ \hline
		\multirow{4}{*}{\begin{tabular}[c]{@{}l@{}}C\&W \\ ($L_2$)\end{tabular}}     & None                                                                            & 100 & 22.46 & 1.972 & 0.514      \\
		& EAD                                                                             & 100 & 26.11 & 2.468 & 0.643      \\
		& C\&W                                                                            & 100 & 24.97 & 2.47  & 0.684      \\
		& EAD + C\&W                                                                      & 100 & 27.32 & 2.513 & 0.653      \\ \hline
		\multirow{4}{*}{\begin{tabular}[c]{@{}l@{}}EAD \\ ($L_1$ rule)\end{tabular}} & None                                                                            & 100 & 14.11 & 2.211 & 0.768      \\
		& EAD                                                                             & 100 & 17.04 & 2.653 & 0.86       \\
		& C\&W                                                                  & 100 & 15.49 & 2.628 & 0.892      \\
		& EAD + C\&W                                                                      & 100 & 16.83 & 2.66  & 0.87       \\ \hline
	\end{tabular}
\end{table}

\subsection{Complementing Adversarial Training}
To further validate the difference between $L_1$-based and $L_2$-based adversarial examples, we test their performance in adversarial training on MNIST. We randomly select 1000 images from the training set and use the C\&W attack and EAD ($L_1$ rule) to generate adversarial examples for all incorrect labels, leading to 9000 adversarial examples in total for each method. We then separately augment the original training set with these examples to retrain the network and test its robustness on the testing set, as summarized in Table \ref{table_adv_training}. For adversarial training with any single method, although both attacks still attain a 100\% success rate in the average case, the network is  more tolerable to adversarial perturbations, as all distortion metrics increase significantly when compared to the null case. 
We also observe that joint adversarial training with EAD and the C\&W method can further increase the $L_1$ and $L_2$ distortions against the  C\&W attack and the $L_2$ distortion against EAD, suggesting that the $L_1$-based examples crafted by EAD can complement adversarial training.

\section{Conclusion}
We proposed an elastic-net regularized attack framework for crafting adversarial examples to attack deep neural networks. Experimental results on MNIST, CIFAR10 and ImageNet show that 
the $L_1$-based adversarial examples crafted by EAD can be as successful as the state-of-the-art $L_2$ and $L_\infty$ attacks in breaking undefended and defensively distilled networks. Furthermore, EAD can improve attack transferability and complement adversarial training. Our results corroborate the effectiveness of EAD and shed new light on the use of $L_1$-based adversarial examples toward adversarial  learning and security implications of deep neural networks.

\textbf{Acknowledgment} Cho-Jui Hsieh and Huan Zhang acknowledge the support of NSF via IIS-1719097.


\bibliography{adversarial_learning}

\begin{thebibliography}{}

\bibitem[\protect\citeauthoryear{Beck and Teboulle}{2009}]{beck2009fast}
Beck, A., and Teboulle, M.
\newblock 2009.
\newblock A fast iterative shrinkage-thresholding algorithm for linear inverse
  problems.
\newblock {\em SIAM journal on imaging sciences} 2(1):183--202.

\bibitem[\protect\citeauthoryear{Cand{\`e}s and
  Wakin}{2008}]{candes2008introduction}
Cand{\`e}s, E.~J., and Wakin, M.~B.
\newblock 2008.
\newblock An introduction to compressive sampling.
\newblock {\em IEEE signal processing magazine} 25(2):21--30.

\bibitem[\protect\citeauthoryear{Carlini and
  Wagner}{2017a}]{carlini2017adversarial}
Carlini, N., and Wagner, D.
\newblock 2017a.
\newblock Adversarial examples are not easily detected: Bypassing ten detection
  methods.
\newblock {\em arXiv preprint arXiv:1705.07263}.

\bibitem[\protect\citeauthoryear{Carlini and
  Wagner}{2017b}]{carlini2017towards}
Carlini, N., and Wagner, D.
\newblock 2017b.
\newblock Towards evaluating the robustness of neural networks.
\newblock In {\em IEEE Symposium on Security and Privacy (SP)},  39--57.

\bibitem[\protect\citeauthoryear{Dong \bgroup et al\mbox.\egroup
  }{2017}]{dong2017towards}
Dong, Y.; Su, H.; Zhu, J.; and Bao, F.
\newblock 2017.
\newblock Towards interpretable deep neural networks by leveraging adversarial
  examples.
\newblock {\em arXiv preprint arXiv:1708.05493}.

\bibitem[\protect\citeauthoryear{Duchi and Singer}{2009}]{duchi2009efficient}
Duchi, J., and Singer, Y.
\newblock 2009.
\newblock Efficient online and batch learning using forward backward splitting.
\newblock {\em Journal of Machine Learning Research} 10(Dec):2899--2934.

\bibitem[\protect\citeauthoryear{Evtimov \bgroup et al\mbox.\egroup
  }{2017}]{Evtimov2017robust}
Evtimov, I.; Eykholt, K.; Fernandes, E.; Kohno, T.; Li, B.; Prakash, A.;
  Rahmati, A.; and Song, D.
\newblock 2017.
\newblock Robust physical-world attacks on machine learning models.
\newblock {\em arXiv preprint arXiv:1707.08945}.

\bibitem[\protect\citeauthoryear{Feinman \bgroup et al\mbox.\egroup
  }{2017}]{feinman2017detecting}
Feinman, R.; Curtin, R.~R.; Shintre, S.; and Gardner, A.~B.
\newblock 2017.
\newblock Detecting adversarial samples from artifacts.
\newblock {\em arXiv preprint arXiv:1703.00410}.

\bibitem[\protect\citeauthoryear{Fu \bgroup et al\mbox.\egroup
  }{2006}]{fu2006efficient}
Fu, H.; Ng, M.~K.; Nikolova, M.; and Barlow, J.~L.
\newblock 2006.
\newblock Efficient minimization methods of mixed l2-l1 and l1-l1 norms for
  image restoration.
\newblock {\em SIAM Journal on scientific computing} 27(6):1881--1902.

\bibitem[\protect\citeauthoryear{Goodfellow, Shlens, and
  Szegedy}{2015}]{goodfellow2014explaining}
Goodfellow, I.~J.; Shlens, J.; and Szegedy, C.
\newblock 2015.
\newblock Explaining and harnessing adversarial examples.
\newblock {\em ICLR'15; arXiv preprint arXiv:1412.6572}.

\bibitem[\protect\citeauthoryear{Grosse \bgroup et al\mbox.\egroup
  }{2017}]{grosse2017statistical}
Grosse, K.; Manoharan, P.; Papernot, N.; Backes, M.; and McDaniel, P.
\newblock 2017.
\newblock On the (statistical) detection of adversarial examples.
\newblock {\em arXiv preprint arXiv:1702.06280}.

\bibitem[\protect\citeauthoryear{Hinton, Vinyals, and
  Dean}{2015}]{hinton2015distilling}
Hinton, G.; Vinyals, O.; and Dean, J.
\newblock 2015.
\newblock Distilling the knowledge in a neural network.
\newblock {\em arXiv preprint arXiv:1503.02531}.

\bibitem[\protect\citeauthoryear{Kingma and Ba}{2014}]{kingma2014adam}
Kingma, D., and Ba, J.
\newblock 2014.
\newblock Adam: A method for stochastic optimization.
\newblock {\em arXiv preprint arXiv:1412.6980}.

\bibitem[\protect\citeauthoryear{Koh and Liang}{2017}]{koh2017understanding}
Koh, P.~W., and Liang, P.
\newblock 2017.
\newblock Understanding black-box predictions via influence functions.
\newblock {\em ICML; arXiv preprint arXiv:1703.04730}.

\bibitem[\protect\citeauthoryear{Kurakin, Goodfellow, and
  Bengio}{2016a}]{kurakin2016adversarial}
Kurakin, A.; Goodfellow, I.; and Bengio, S.
\newblock 2016a.
\newblock Adversarial examples in the physical world.
\newblock {\em arXiv preprint arXiv:1607.02533}.

\bibitem[\protect\citeauthoryear{Kurakin, Goodfellow, and
  Bengio}{2016b}]{kurakin2016adversarial_ICLR}
Kurakin, A.; Goodfellow, I.; and Bengio, S.
\newblock 2016b.
\newblock Adversarial machine learning at scale.
\newblock {\em ICLR'17; arXiv preprint arXiv:1611.01236}.

\bibitem[\protect\citeauthoryear{Liu \bgroup et al\mbox.\egroup
  }{2016}]{liu2016delving}
Liu, Y.; Chen, X.; Liu, C.; and Song, D.
\newblock 2016.
\newblock Delving into transferable adversarial examples and black-box attacks.
\newblock {\em arXiv preprint arXiv:1611.02770}.

\bibitem[\protect\citeauthoryear{Lu, Issaranon, and
  Forsyth}{2017}]{lu2017safetynet}
Lu, J.; Issaranon, T.; and Forsyth, D.
\newblock 2017.
\newblock Safetynet: Detecting and rejecting adversarial examples robustly.
\newblock {\em arXiv preprint arXiv:1704.00103}.

\bibitem[\protect\citeauthoryear{Madry \bgroup et al\mbox.\egroup
  }{2017}]{madry2017towards}
Madry, A.; Makelov, A.; Schmidt, L.; Tsipras, D.; and Vladu, A.
\newblock 2017.
\newblock Towards deep learning models resistant to adversarial attacks.
\newblock {\em arXiv preprint arXiv:1706.06083}.

\bibitem[\protect\citeauthoryear{Moosavi-Dezfooli \bgroup et al\mbox.\egroup
  }{2016}]{moosavi2016universal}
Moosavi-Dezfooli, S.-M.; Fawzi, A.; Fawzi, O.; and Frossard, P.
\newblock 2016.
\newblock Universal adversarial perturbations.
\newblock {\em arXiv preprint arXiv:1610.08401}.

\bibitem[\protect\citeauthoryear{Moosavi-Dezfooli, Fawzi, and
  Frossard}{2016}]{moosavi2016deepfool}
Moosavi-Dezfooli, S.-M.; Fawzi, A.; and Frossard, P.
\newblock 2016.
\newblock Deepfool: a simple and accurate method to fool deep neural networks.
\newblock In {\em Proceedings of the IEEE Conference on Computer Vision and
  Pattern Recognition},  2574--2582.

\bibitem[\protect\citeauthoryear{Papernot \bgroup et al\mbox.\egroup
  }{2016a}]{papernot2016limitations}
Papernot, N.; McDaniel, P.; Jha, S.; Fredrikson, M.; Celik, Z.~B.; and Swami,
  A.
\newblock 2016a.
\newblock The limitations of deep learning in adversarial settings.
\newblock In {\em IEEE European Symposium on Security and Privacy (EuroS\&P)},
  372--387.

\bibitem[\protect\citeauthoryear{Papernot \bgroup et al\mbox.\egroup
  }{2016b}]{papernot2016distillation}
Papernot, N.; McDaniel, P.; Wu, X.; Jha, S.; and Swami, A.
\newblock 2016b.
\newblock Distillation as a defense to adversarial perturbations against deep
  neural networks.
\newblock In {\em IEEE Symposium on Security and Privacy (SP)},  582--597.

\bibitem[\protect\citeauthoryear{Papernot \bgroup et al\mbox.\egroup
  }{2017}]{papernot2017practical}
Papernot, N.; McDaniel, P.; Goodfellow, I.; Jha, S.; Celik, Z.~B.; and Swami,
  A.
\newblock 2017.
\newblock Practical black-box attacks against machine learning.
\newblock In {\em ACM Asia Conference on Computer and Communications Security},
   506--519.

\bibitem[\protect\citeauthoryear{Parikh, Boyd, and
  others}{2014}]{parikh2014proximal}
Parikh, N.; Boyd, S.; et~al.
\newblock 2014.
\newblock Proximal algorithms.
\newblock {\em Foundations and Trends{\textregistered} in Optimization}
  1(3):127--239.

\bibitem[\protect\citeauthoryear{Szegedy \bgroup et al\mbox.\egroup
  }{2013}]{szegedy2013intriguing}
Szegedy, C.; Zaremba, W.; Sutskever, I.; Bruna, J.; Erhan, D.; Goodfellow, I.;
  and Fergus, R.
\newblock 2013.
\newblock Intriguing properties of neural networks.
\newblock {\em arXiv preprint arXiv:1312.6199}.

\bibitem[\protect\citeauthoryear{Szegedy \bgroup et al\mbox.\egroup
  }{2016}]{szegedy2016rethinking}
Szegedy, C.; Vanhoucke, V.; Ioffe, S.; Shlens, J.; and Wojna, Z.
\newblock 2016.
\newblock Rethinking the inception architecture for computer vision.
\newblock In {\em Proceedings of the IEEE Conference on Computer Vision and
  Pattern Recognition},  2818--2826.

\bibitem[\protect\citeauthoryear{Tram{\`e}r \bgroup et al\mbox.\egroup
  }{2017}]{tramer2017ensemble}
Tram{\`e}r, F.; Kurakin, A.; Papernot, N.; Boneh, D.; and McDaniel, P.
\newblock 2017.
\newblock Ensemble adversarial training: Attacks and defenses.
\newblock {\em arXiv preprint arXiv:1705.07204}.

\bibitem[\protect\citeauthoryear{Xu, Evans, and Qi}{2017}]{xu2017feature}
Xu, W.; Evans, D.; and Qi, Y.
\newblock 2017.
\newblock Feature squeezing: Detecting adversarial examples in deep neural
  networks.
\newblock {\em arXiv preprint arXiv:1704.01155}.

\bibitem[\protect\citeauthoryear{Zantedeschi, Nicolae, and
  Rawat}{2017}]{zantedeschi2017efficient}
Zantedeschi, V.; Nicolae, M.-I.; and Rawat, A.
\newblock 2017.
\newblock Efficient defenses against adversarial attacks.
\newblock {\em arXiv preprint arXiv:1707.06728}.

\bibitem[\protect\citeauthoryear{Zheng \bgroup et al\mbox.\egroup
  }{2016}]{zheng2016improving}
Zheng, S.; Song, Y.; Leung, T.; and Goodfellow, I.
\newblock 2016.
\newblock Improving the robustness of deep neural networks via stability
  training.
\newblock In {\em Proceedings of the IEEE Conference on Computer Vision and
  Pattern Recognition},  4480--4488.

\bibitem[\protect\citeauthoryear{Zou and Hastie}{2005}]{zou2005regularization}
Zou, H., and Hastie, T.
\newblock 2005.
\newblock Regularization and variable selection via the elastic net.
\newblock {\em Journal of the Royal Statistical Society: Series B (Statistical
  Methodology)} 67(2):301--320.

\end{thebibliography}
\bibliographystyle{aaai}

 \clearpage
 \section{Supplementary Material}

 \subsection{Proof of Optimality of (\ref{eqn_ISTA}) for Solving EAD in (\ref{eqn_elastic_net})}
 Since the $L_1$ penalty $\beta \|\bx-\bx_0\|_1$ in (\ref{eqn_elastic_net}) is a non-differentiable yet smooth function, we use the proximal gradient method \cite{parikh2014proximal} for solving the EAD formulation in (\ref{eqn_elastic_net}). Define $\Phi_{\cZ}(\bz)$ to be the indicator function of an interval $\cZ$ such that $\Phi_{\cZ}(\bz)=0$ if $\bz \in \cZ$ and $\Phi_{\cZ}(\bz)=\infty$ if $\bz \notin \cZ$. Using $\Phi_{\cZ}(\bz)$, the EAD formulation in (\ref{eqn_elastic_net}) can be rewritten as
 \begin{align}
 \label{eqn_EAD_formulation_prox}
 \textnormal{minimize}_{\bx \in \bbR^p}~~ g(\bx) + \beta \|\bx-\bx_0\|_1 + \Phi_{[0,1]^p}(\bx),
 \end{align}
 where $g(\bx)=c \cdot f(\bx,t)+  \|\bx- \bx_0\|_2^2 $. The proximal operator $\textnormal{Prox}(\bx)$ of $\beta \|\bx-\bx_0\|_1$ constrained to $\bx \in [0,1]^p$ is 
 \begin{align}
 \label{eqn_prox}
 \textnormal{Prox}(\bx)&=\arg \min _{\bz \in \bbR^p} \frac{1}{2}\|\bz -  \bx\|_2^2+ \beta \|\bz-\bx_0\|_1 + \Phi_{[0,1]^p}(\bz)  \nonumber \\
 &=\arg \min_{\bz \in [0,1]^p} \frac{1}{2}\|\bz - \bx\|_2^2+ \beta \|\bz-\bx_0\|_1 \nonumber \\
 &= S_{\beta}(\bx),
 \end{align}
 where the mapping function $S_{\beta}$ is defined in (\ref{eqn_ISTA_S}). Consequently, using (\ref{eqn_prox}), the proximal gradient algorithm for solving (\ref{eqn_EAD_formulation}) is iterated by
 \begin{align}
 \label{eqn_prox_2}
 \bx^{(k+1)}&=\textnormal{Prox} (\bx^{(k)} - \alpha_k \nabla g(\bx^{(k)})) \\
 &= S_\beta (\bx^{(k)} - \alpha_k \nabla g(\bx^{(k)})),
 \end{align}
 which completes the proof.

 \subsection{Grid Search for FGM and I-FGM (Table \ref{Table_grid_search})}
 To determine the optimal distortion parameter $\epsilon$ for FGM and I-FGM methods, we adopt a fine grid search on $\epsilon$. For each image, the best parameter is the smallest $\epsilon$ in the grid leading to a successful targeted attack. If the grid search fails to find a successful adversarial example, the attack is considered in vain. The selected range for grid search covers the reported distortion statistics of EAD and the C\&W attack. 
 The resolution of the grid search for FGM is selected such that it will generate 1000 candidates of adversarial examples during the grid search per input image. The resolution of the grid search for I-FGM is selected such that it will compute gradients  for 10000 times in total (i.e., 1000 FGM operations $\times$ 10 iterations) during the grid search per input image, which is more than the total number of gradients (9000)  computed by EAD and the C\&W attack.

 \begin{table}[h]
 	\centering
 	\caption{Range and resolution of grid search  for finding the optimal distortion parameter $\epsilon$ for FGM and I-FGM.}
 	\label{Table_grid_search}
 \begin{tabular}{l|ll}
 	\hline
 	\multicolumn{1}{c|}{}       & \multicolumn{2}{c}{Grid search}                            \\ \hline
 	\multicolumn{1}{c|}{Method} & \multicolumn{1}{c}{Range} & \multicolumn{1}{c}{Resolution} \\ \hline
 	FGM-$L_\infty$              & $[10^{-3},1]$             & $10^{-3}$                      \\
 	FGM-$L_1$                   & $[1,10^3]$                & $1$                            \\
 	FGM-$L_2$                   & $[10^{-2},10]$            & $10^{-2}$                      \\
 	I-FGM-$L_\infty$            & $[10^{-3},1]$             & $10^{-3}$                      \\
 	I-FGM-$L_1$                 & $[1,10^3]$                & $1$                            \\
 	I-FGM-$L_2$                 & $[10^{-2},10]$            & $10^{-2}$                      \\ \hline
 \end{tabular}
 \end{table}

 \subsection{Comparison of COV and EAD on CIFAR10 (Table \ref{table_COV_EAD_cifar})} 

 Table \ref{table_COV_EAD_cifar} compares the attack performance of using EAD (Algorithm \ref{algo_EAD})) and the change-of-variable (COV) approach for solving the elastic-net formulation in (\ref{eqn_EAD_formulation}) on CIFAR10.  Similar to the MNIST results in Table \ref{table_COV_EAD}, although COV and EAD attain similar attack success rates, we find that COV is not effective in crafting $L_1$-based adversarial examples. Increasing  $\beta$ leads to less $L_1$-distorted adversarial examples for EAD, whereas the distortion ($L_1$, $L_2$ and $L_\infty$) of COV is insensitive to changes in $\beta$. The insensitivity of COV suggests that it is inadequate for elastic-net optimization, which can be explained by its inefficiency in subgradient-based optimization problems \cite{duchi2009efficient}.  

 \begin{table*}[]
 	\centering
 	\caption{Comparison of the change-of-variable (COV) approach and EAD (Algorithm \ref{algo_EAD}) for solving the elastic-net formulation in (\ref{eqn_EAD_formulation}) on CIFAR10.  ASR means attack success rate (\%). Similar to the results on MNIST in Table \ref{table_COV_EAD}, increasing $\beta$ leads to less $L_1$-distorted adversarial examples for EAD, whereas the distortion of COV is insensitive to changes in $\beta$.}
 	\label{table_COV_EAD_cifar}
 \begin{tabular}{ll|llll|llll|llll}
 	\hline
 	&           & \multicolumn{4}{c|}{Best case}   & \multicolumn{4}{c|}{Average case}      & \multicolumn{4}{c}{Worst case}    \\ \hline
 	\begin{tabular}[c]{@{}l@{}}Optimization \\ method\end{tabular}           & $\beta$   & ASR & $L_1$ & $L_2$ & $L_\infty$ & ASR  & $L_1$ & $L_\infty$ & $L_\infty$ & ASR  & $L_1$ & $L_2$ & $L_\infty$ \\ \hline
 	\multirow{5}{*}{COV}                                                     & $0$       & 100 & 7.167 & 0.209 & 0.022      & 100  & 13.62 & 0.392      & 0.044      & 99.9 & 19.17 & 0.546 & 0.064      \\
 	& $10^{-5}$ & 100 & 7.165 & 0.209 & 0.022      & 100  & 13.43 & 0.386      & 0.044      & 100  & 19.16 & 0.546 & 0.065      \\
 	& $10^{-4}$ & 100 & 7.148 & 0.209 & 0.022      & 99.9 & 13.64 & 0.394      & 0.044      & 99.9 & 19.14 & 0.547 & 0.064      \\
 	& $10^{-3}$ & 100 & 6.961 & 0.211 & 0.023      & 100  & 13.35 & 0.396      & 0.045      & 100  & 18.7  & 0.547 & 0.066      \\
 	& $10^{-2}$ & 100 & 5.963 & 0.222 & 0.029      & 100  & 11.51 & 0.408      & 0.055      & 100  & 16.31 & 0.556 & 0.077      \\ \hline
 	\multirow{5}{*}{\begin{tabular}[c]{@{}l@{}}EAD\\ (EN rule)\end{tabular}} & $0$       & 100 & 6.643 & 0.201 & 0.023      & 99.9 & 13.39 & 0.392      & 0.045      & 99.9 & 18.72 & 0.541 & 0.064      \\
 	& $10^{-5}$ & 100 & 5.967 & 0.201 & 0.025      & 100  & 12.24 & 0.391      & 0.047      & 99.7 & 17.24 & 0.541 & 0.068      \\
 	& $10^{-4}$ & 100 & 4.638 & 0.215 & 0.032      & 99.9 & 10.4  & 0.414      & 0.058      & 99.9 & 14.86 & 0.56  & 0.081      \\
 	& $10^{-3}$ & 100 & 4.014 & 0.261 & 0.047      & 100  & 8.18  & 0.502      & 0.097      & 100  & 12.11 & 0.69  & 0.147      \\
 	& $10^{-2}$ & 100 & 3.688 & 0.357 & 0.085      & 100  & 8.106 & 0.741      & 0.209      & 100  & 12.59 & 1.053 & 0.351      \\ \hline
 \end{tabular}
 \end{table*}

 \subsection{Comparison of EN and $L_1$ decision rules in EAD on MNIST and CIFAR10 (Tables \ref{table_rule_MNIST} and  \ref{table_rule_cifar})} 

\begin{figure*}[t]
 	\centering
 	\begin{subfigure}[b]{0.25\linewidth}
 \includegraphics[width=\textwidth]{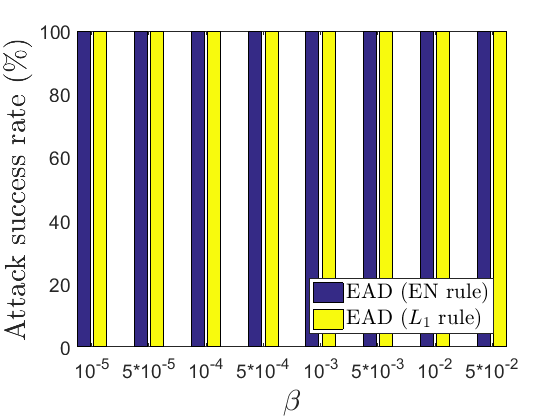}
 	\end{subfigure}%
 	\centering
 	\begin{subfigure}[b]{0.25\linewidth}
 		\includegraphics[width=\textwidth]{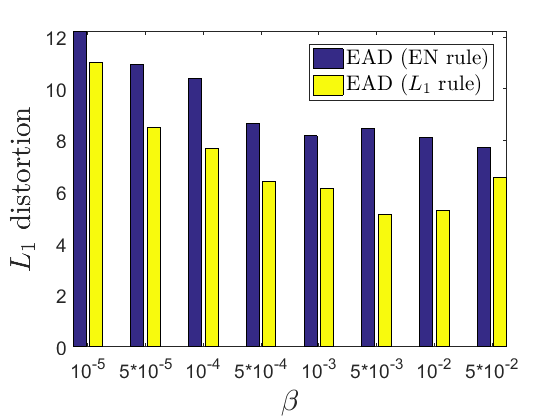}
 	\end{subfigure}%
 	\centering
 	\begin{subfigure}[b]{0.25\linewidth}
 		\includegraphics[width=\textwidth]{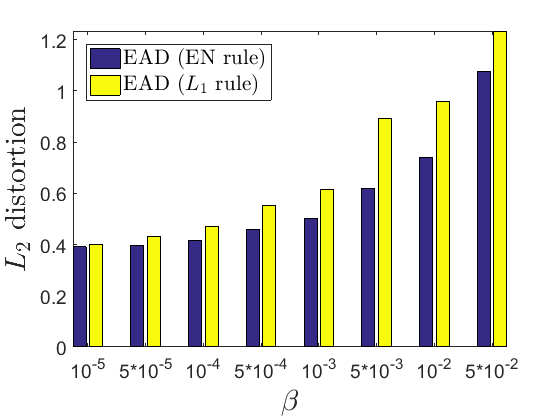}
 	\end{subfigure}%
 	\centering
 	\begin{subfigure}[b]{0.25\linewidth}
 		\includegraphics[width=\textwidth]{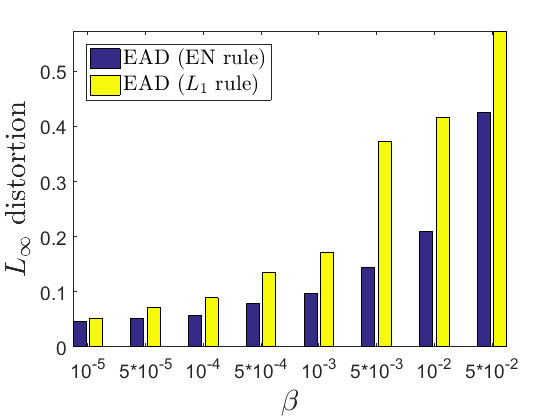}
 	\end{subfigure}
 	\caption{Comparison of EN and $L_1$ decision rules in EAD on CIFAR10 with varying $L_1$ regularization parameter $\beta$ (average case). Comparing to the EN rule, for the same $\beta$
 		the $L_1$ rule attains less $L_1$ distortion but may incur more  $L_2$ and $L_\infty$ distortions.  }
 	\label{Fig_rule_cifar}
 \end{figure*}

 Figure \ref{Fig_rule_cifar} compares the average-case distortion of these two decision rules with different values of $\beta$ on CIFAR10. For the same $\beta$, the $L_1$ rule gives less $L_1$ distorted adversarial examples than those given by the EN rule at the price of larger $L_2$ and $L_\infty$ distortions. We also observe that the $L_1$ distortion does not decrease monotonically with $\beta$. In particular,
 large $\beta$ values  (e.g., $\beta=5 \cdot 10^{-2}$) may lead to increased $L_1$ distortion due to excessive shrinking and thresholding. Table \ref{table_rule_MNIST} and Table \ref{table_rule_cifar} displays the complete attack results of these two decision rules on MNIST and CIFAR10, respectively.

 \begin{table*}[t]
 	\centering
 	\caption{Comparison of the elastic net (EN) and $L_1$ decision rules in EAD for selecting adversarial examples on MNIST. ASR means attack success rate (\%).}
 	\label{table_rule_MNIST}
 \begin{tabular}{ll|llll|llll|llll}
 	\hline
 	&                   & \multicolumn{4}{c|}{Best case}   & \multicolumn{4}{c|}{Average case} & \multicolumn{4}{c}{Worst case}    \\ \hline
 	\begin{tabular}[c]{@{}l@{}}Decision\\ rule\end{tabular}                     & $\beta$           & ASR & $L_1$ & $L_2$ & $L_\infty$ & ASR  & $L_1$ & $L_2$ & $L_\infty$ & ASR  & $L_1$ & $L_2$ & $L_\infty$ \\ \hline
 	\multirow{8}{*}{\begin{tabular}[c]{@{}l@{}}EAD\\ (EN rule)\end{tabular}}    & $10^{-5}$         & 100 & 13.66 & 1.369 & 0.378      & 100  & 22.6  & 1.98  & 0.515      & 99.9 & 30.79 & 2.507 & 0.648      \\
 	& $5 \cdot 10^{-5}$ & 100 & 13.19 & 1.37  & 0.383      & 100  & 21.25 & 1.947 & 0.518      & 100  & 29.82 & 2.512 & 0.659      \\
 	& $10^{-4}$         & 100 & 12.79 & 1.372 & 0.388      & 100  & 20.98 & 1.951 & 0.521      & 100  & 29.21 & 2.514 & 0.667      \\
 	& $5 \cdot 10^{-4}$ & 100 & 10.95 & 1.395 & 0.42       & 100  & 19.11 & 1.986 & 0.558      & 100  & 27.04 & 2.544 & 0.706      \\
 	& $ 10^{-3}$        & 100 & 9.808 & 1.427 & 0.452      & 100  & 17.4  & 2.001 & 0.594      & 100  & 25.52 & 2.582 & 0.748      \\
 	& $5 \cdot 10^{-3}$ & 100 & 7.912 & 1.6   & 0.591      & 100  & 14.49 & 2.261 & 0.772      & 100  & 21.64 & 2.835 & 0.921      \\
 	& $10^{-2}$         & 100 & 7.271 & 1.718 & 0.674      & 100  & 13.56 & 2.395 & 0.852      & 100  & 20.77 & 3.021 & 0.976      \\
 	& $5 \cdot 10^{-2}$ & 100 & 7.088 & 1.872 & 0.736      & 100  & 13.38 & 2.712 & 0.919      & 100  & 20.19 & 3.471 & 0.998      \\ \hline
 	\multirow{8}{*}{\begin{tabular}[c]{@{}l@{}}EAD\\ ($L_1$ rule)\end{tabular}} & $10^{-5}$         & 100 & 13.26 & 1.376 & 0.386      & 100  & 21.75 & 1.965 & 0.525      & 100  & 30.48 & 2.521 & 0.666      \\
 	& $5 \cdot 10^{-4}$ & 100 & 11.81 & 1.385 & 0.404      & 100  & 20.28 & 1.98  & 0.54       & 100  & 28.8  & 2.527 & 0.686      \\
 	& $10^{-4}$         & 100 & 10.75 & 1.403 & 0.424      & 100  & 18.69 & 1.983 & 0.565      & 100  & 27.49 & 2.539 & 0.71       \\
 	& $5 \cdot 10^{-4}$ & 100 & 8.025 & 1.534 & 0.527      & 100  & 15.42 & 2.133 & 0.694      & 100  & 23.62 & 2.646 & 0.857      \\
 	& $10^{-3}$         & 100 & 7.153 & 1.639 & 0.593      & 100  & 14.11 & 2.211 & 0.768      & 100  & 22.05 & 2.747 & 0.934      \\
 	& $5 \cdot 10^{-3}$ & 100 & 6.347 & 1.844 & 0.772      & 100  & 11.99 & 2.491 & 0.927      & 100  & 18.01 & 3.218 & 0.997      \\
 	& $10^{-2}$         & 100 & 6.193 & 1.906 & 0.861      & 100  & 11.69 & 2.68  & 0.97       & 100  & 17.29 & 3.381 & 1          \\
 	& $5 \cdot 10^{-2}$ & 100 & 5.918 & 2.101 & 0.956      & 100  & 11.59 & 2.873 & 0.993      & 100  & 18.67 & 3.603 & 1          \\ \hline
 \end{tabular}
 \end{table*}

 \begin{table*}[t]
 	\centering
 	\caption{Comparison of the elastic net (EN) and $L_1$ decision rules in EAD for selecting adversarial examples on CIFAR10. ASR means attack success rate (\%).}
 	\label{table_rule_cifar}
 \begin{tabular}{ll|llllllll|llll}
 	\hline
 	&                   & \multicolumn{4}{c}{Best case}                         & \multicolumn{4}{c|}{Average case} & \multicolumn{4}{c}{Worst case}    \\ \hline
 	\begin{tabular}[c]{@{}l@{}}Decision \\ rule\end{tabular}                    & $\beta$           & ASR & $L_1$  & $L_2$  & \multicolumn{1}{l|}{$L_\infty$} & ASR  & $L_1$  & $L_2$  & $L_\infty$ & ASR  & $L_1$  & $L_2$  & $L_\infty$ \\ \hline
 	\multirow{8}{*}{\begin{tabular}[c]{@{}l@{}}EAD\\ (EN rule)\end{tabular}}    & $10^{-5}$         & 100 & 5.967 & 0.201 & \multicolumn{1}{l|}{0.025}      & 100  & 12.24 & 0.391 & 0.047      & 99.7 & 17.24 & 0.541 & 0.068      \\
 	& $5 \cdot 10^{-5}$ & 100 & 5.09  & 0.207 & \multicolumn{1}{l|}{0.028}      & 99.9 & 10.94 & 0.396 & 0.052      & 99.9 & 15.87 & 0.549 & 0.075      \\
 	& $10^{-4}$         & 100 & 4.638 & 0.215 & \multicolumn{1}{l|}{0.032}      & 99.9 & 10.4  & 0.414 & 0.058      & 99.9 & 14.86 & 0.56  & 0.081      \\
 	& $5 \cdot 10^{-4}$ & 100 & 4.35  & 0.241 & \multicolumn{1}{l|}{0.039}      & 100  & 8.671 & 0.459 & 0.079      & 100  & 12.39 & 0.63  & 0.117      \\
 	& $10^{-3}$         & 100 & 4.014 & 0.261 & \multicolumn{1}{l|}{0.047}      & 100  & 8.18  & 0.502 & 0.097      & 100  & 12.11 & 0.69  & 0.147      \\
 	& $5 \cdot 10^{-3}$ & 100 & 3.83  & 0.319 & 0.067                           & 100  & 8.462 & 0.619 & 0.145      & 100  & 13.03 & 0.904 & 0.255      \\
 	& $10^{-2}$         & 100 & 3.688 & 0.357 & \multicolumn{1}{l|}{0.085}      & 100  & 8.106 & 0.741 & 0.209      & 100  & 12.59 & 1.053 & 0.351      \\
 	& $5 \cdot 10^{-2}$ & 100 & 2.855 & 0.52  & \multicolumn{1}{l|}{0.201}      & 100  & 7.735 & 1.075 & 0.426      & 100  & 14.66 & 1.657 & 0.661      \\ \hline
 	\multirow{8}{*}{\begin{tabular}[c]{@{}l@{}}EAD\\ ($L_1$ rule)\end{tabular}} & $10^{-5}$         & 100 & 5.002 & 0.209 & \multicolumn{1}{l|}{0.029}      & 100  & 11.03 & 0.4   & 0.052      & 99.8 & 16.03 & 0.548 & 0.074      \\
 	& $5 \cdot 10^{-5}$ & 100 & 3.884 & 0.231 & \multicolumn{1}{l|}{0.04}       & 100  & 8.516 & 0.431 & 0.071      & 100  & 12.91 & 0.585 & 0.099      \\
 	& $10^{-4}$         & 100 & 3.361 & 0.255 & \multicolumn{1}{l|}{0.05}       & 100  & 7.7   & 0.472 & 0.089      & 100  & 11.7  & 0.619 & 0.121      \\
 	& $5 \cdot 10^{-4}$ & 100 & 2.689 & 0.339 & \multicolumn{1}{l|}{0.091}      & 100  & 6.414 & 0.552 & 0.135      & 100  & 10.11 & 0.732 & 0.192      \\
 	& $10^{-3}$         & 100 & 2.6   & 0.358 & \multicolumn{1}{l|}{0.103}      & 100  & 6.127 & 0.617 & 0.171      & 100  & 8.99  & 0.874 & 0.272      \\
 	& $5 \cdot 10^{-3}$ & 100 & 2.216 & 0.521 & \multicolumn{1}{l|}{0.22}       & 100  & 5.15  & 0.894 & 0.372      & 100  & 7.983 & 1.195 & 0.542      \\
 	& $10^{-2}$         & 100 & 2.201 & 0.568 & \multicolumn{1}{l|}{0.256}      & 100  & 5.282 & 0.958 & 0.417      & 100  & 8.437 & 1.274 & 0.593      \\
 	& $5 \cdot 10^{-2}$ & 100 & 2.306 & 0.674 & \multicolumn{1}{l|}{0.348}      & 100  & 6.566 & 1.234 & 0.573      & 100  & 12.81 & 1.804 & 0.779      \\ \hline
 \end{tabular}
 \end{table*}

\begin{figure*}[h]
 		\centering
 		\begin{subfigure}[b]{0.3\linewidth}
 			\includegraphics[width=\textwidth]{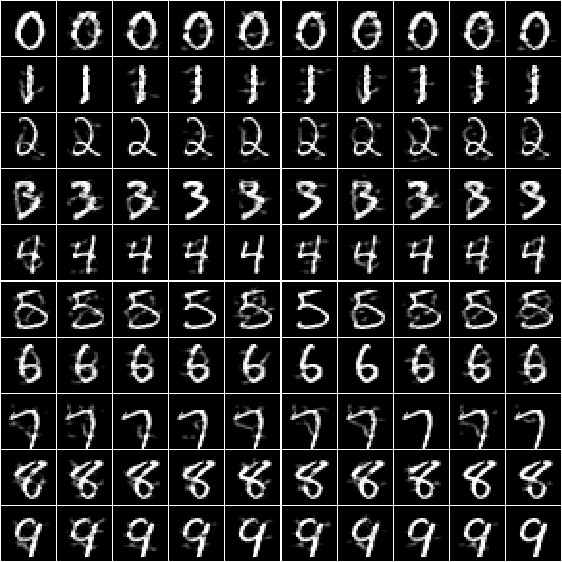}
 			\caption{EAD (EN rule)}
 		\end{subfigure}%
 		\hspace{0.5cm}
 		\centering
 		\begin{subfigure}[b]{0.3\linewidth}
 			\includegraphics[width=\textwidth]{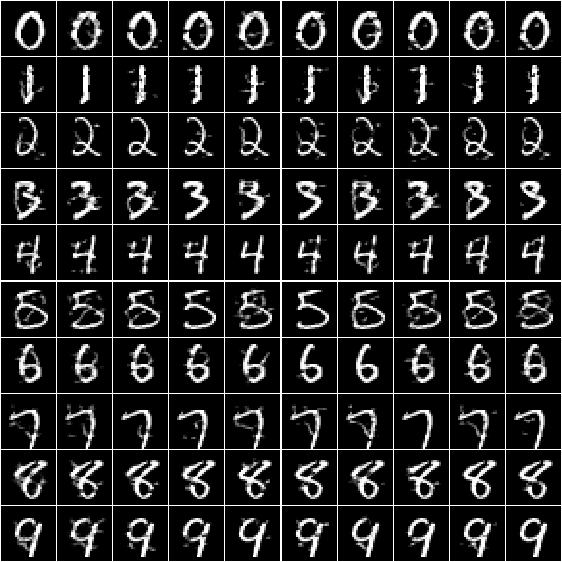}
 			\caption{EAD ($L_1$ rule)}
 		\end{subfigure}%
 		\hspace{0.5cm}				
 		\centering
 		\begin{subfigure}[b]{0.3\linewidth}
 			\includegraphics[width=\textwidth]{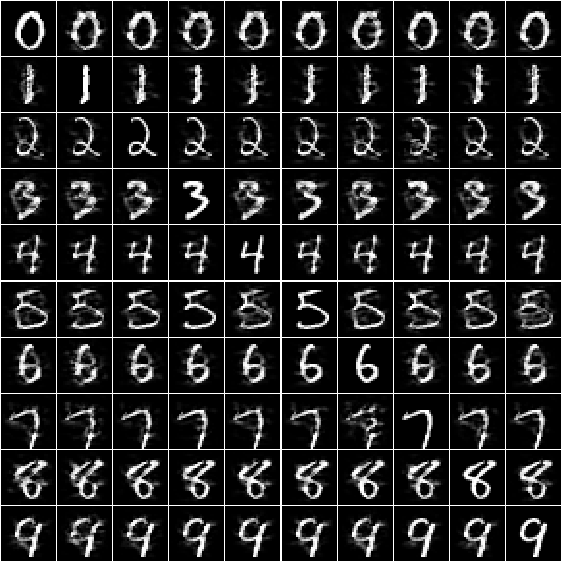}
 			\caption{I-FGM-$L_1$}
 		\end{subfigure}
 		\\
 				\vspace{0.5cm}
 		\centering
 		\begin{subfigure}[b]{0.3\linewidth}
 			\includegraphics[width=\textwidth]{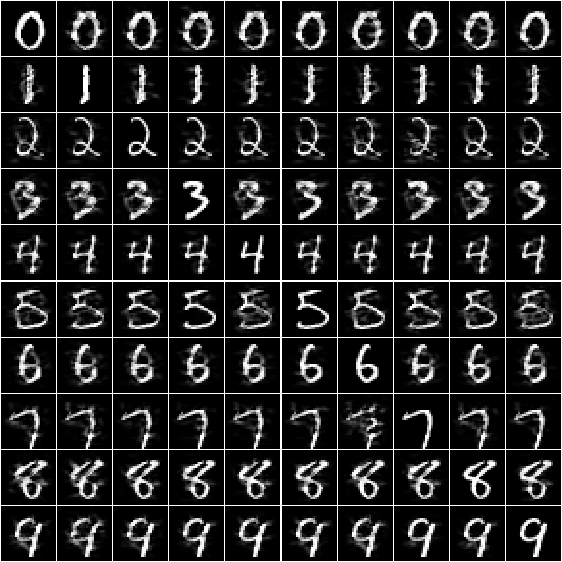}
 			\caption{I-FGM-$L_2$}
 		\end{subfigure}%
 		\hspace{0.5cm}
 		\centering
 		\begin{subfigure}[b]{0.3\linewidth}
 			\includegraphics[width=\textwidth]{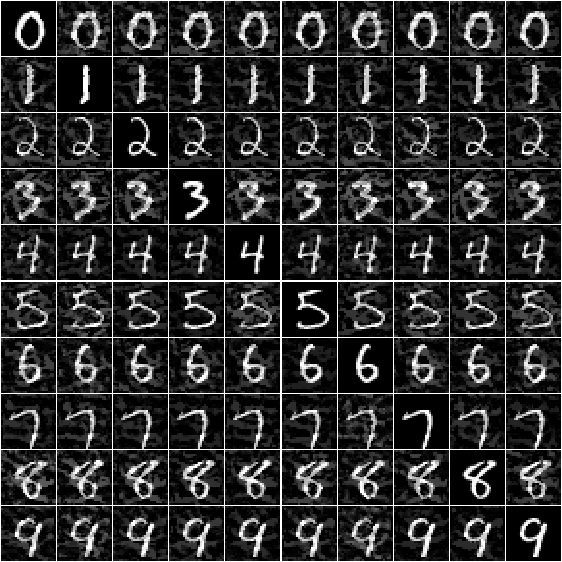}
 			\caption{I-FGM-$L_\infty$}
 		\end{subfigure}%
 		\hspace{0.5cm}				
 		\centering
 		\begin{subfigure}[b]{0.3\linewidth}
 			\includegraphics[width=\textwidth]{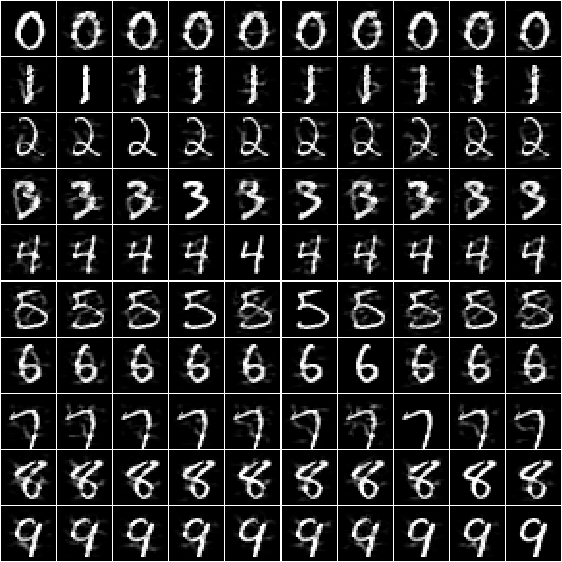}
 			\caption{C\&W}
 		\end{subfigure}		
 		\caption{Visual illustration of adversarial examples crafted by different attack methods on MNIST. For each method, the images displayed on the diagonal are the original examples. In each row, the off-diagonal images are the corresponding adversarial examples with columns indexing target labels (from left to right: digits 0 to 9).}
 		\label{Fig_mnist_example}
 	\end{figure*}

\begin{figure*}[h]
 		\centering
 		\begin{subfigure}[b]{0.3\linewidth}
 			\includegraphics[width=\textwidth]{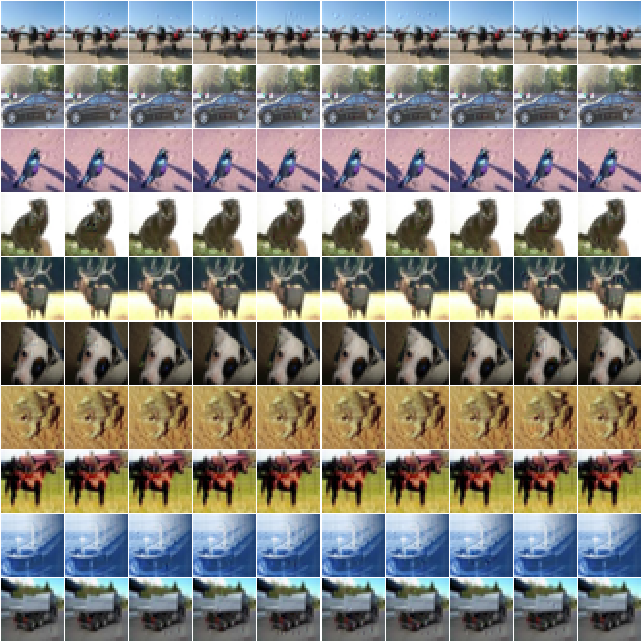}
 			\caption{EAD (EN rule)}
 		\end{subfigure}%
 		\hspace{0.5cm}
 		\centering
 		\begin{subfigure}[b]{0.3\linewidth}
 			\includegraphics[width=\textwidth]{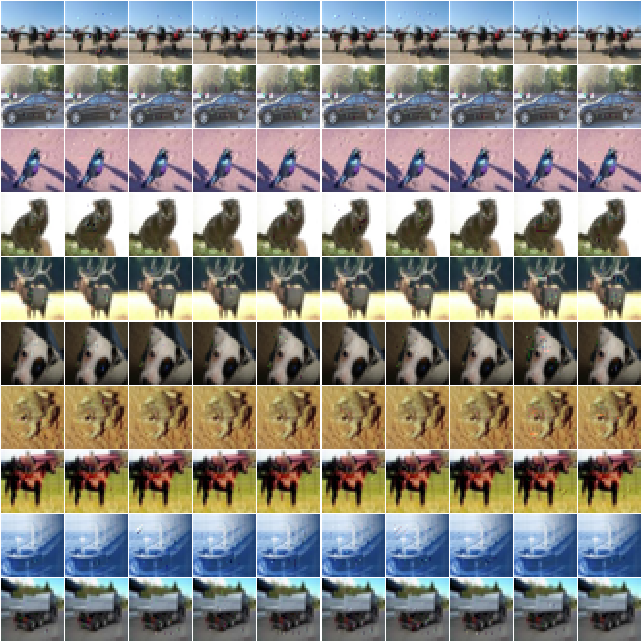}
 			\caption{EAD ($L_1$ rule)}
 		\end{subfigure}%
 		\hspace{0.5cm}				
 		\centering
 		\begin{subfigure}[b]{0.3\linewidth}
 			\includegraphics[width=\textwidth]{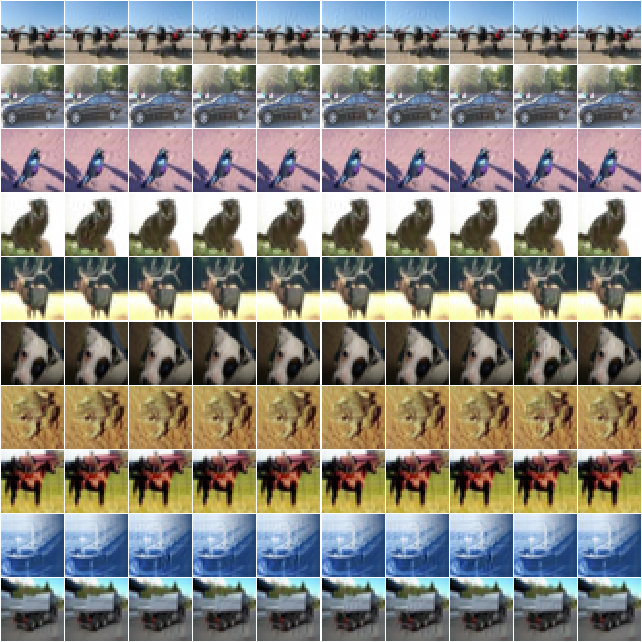}
 			\caption{I-FGM-$L_1$}
 		\end{subfigure}
 		\\
 		\vspace{0.5cm}
 		\centering
 		\begin{subfigure}[b]{0.3\linewidth}
 			\includegraphics[width=\textwidth]{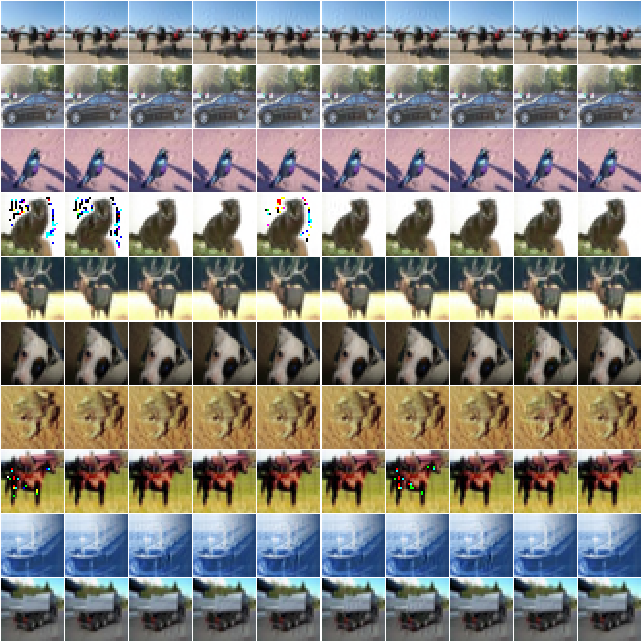}
 			\caption{I-FGM-$L_2$}
 		\end{subfigure}%
 		\hspace{0.5cm}
 		\centering
 		\begin{subfigure}[b]{0.3\linewidth}
 			\includegraphics[width=\textwidth]{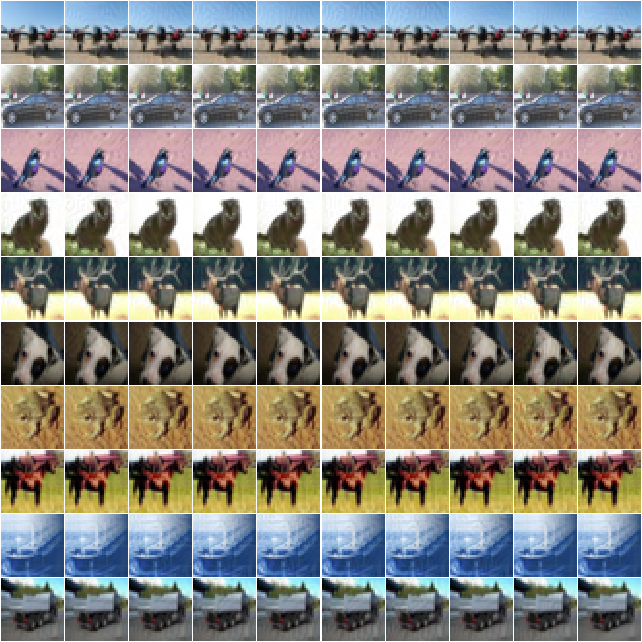}
 			\caption{I-FGM-$L_\infty$}
 		\end{subfigure}%
 		\hspace{0.5cm}				
 		\centering
 		\begin{subfigure}[b]{0.3\linewidth}
 			\includegraphics[width=\textwidth]{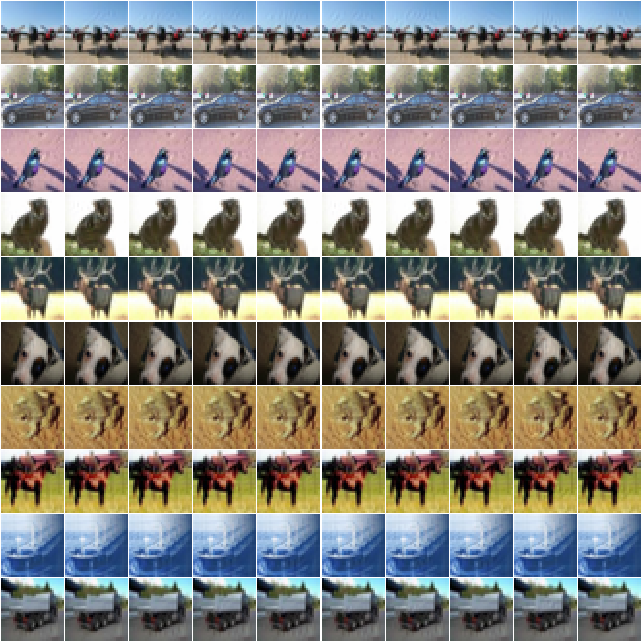}
 			\caption{C\&W}
 		\end{subfigure}		
 		\caption{Visual illustration of adversarial examples crafted by different attack  methods on CIFAR10. For each method, the images displayed on the diagonal are the original examples. In each row, the off-diagonal images are the corresponding adversarial examples with columns indexing target labels.}
 				\label{Fig_cifar_example}
 	\end{figure*}

\begin{figure*}[h]
 	\centering
 	\includegraphics[width=1\linewidth]{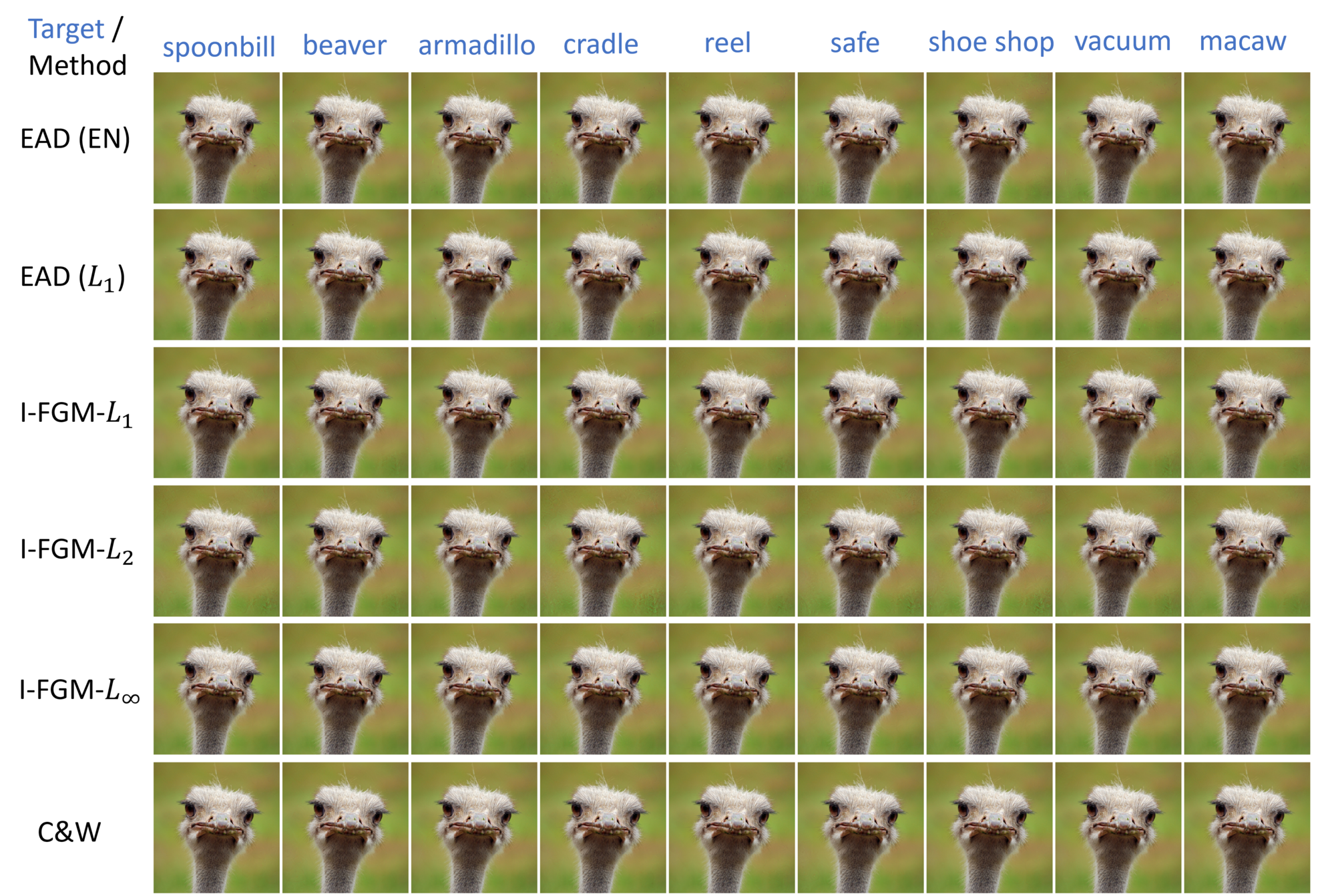}
 	\caption{Visual illustration of adversarial examples crafted by different attack  methods on ImageNet. The original example is the ostrich image (Figure \ref{Fig_ostrich_demo} (a)). Each column represents a targeted class to attack, and each row represents an attack method. 
 		}.
 	\label{Fig_ostrich_all_attack}
 \end{figure*}

 \subsection{Complete Attack Results and Visual Illustration on MNIST, CIFAR10 and ImageNet (Tables \ref{table_attack_mnist},  \ref{table_attack_cifar} and \ref{table_attack_imagenet} and Figures \ref{Fig_mnist_example}, \ref{Fig_cifar_example} and \ref{Fig_ostrich_all_attack})}

 Tables \ref{table_attack_mnist},  \ref{table_attack_cifar} and \ref{table_attack_imagenet} summarize the complete attack results of all the considered attack methods on MNIST, CIFAR10 and ImageNet, respectively. EAD, the C\&W attack and I-FGM all lead to 100\% attack success rate in the average case.
 Among the three image classification datasets, ImageNet is the easiest one to attack due to low distortion per image pixel, and MNIST is the most difficult one to attack. 
 For the purpose of visual illustration, the adversarial examples of selected benign images from the test sets are displayed in  Figures \ref{Fig_mnist_example}, \ref{Fig_cifar_example} and \ref{Fig_ostrich_all_attack}. On CIFAR10 and ImageNet, the adversarial examples are visually indistinguishable. On MNIST, the I-FGM examples are blurrier than EAD and the C\&W attack.

 \begin{table*}[]
 	\centering
 	\caption{Comparison of different adversarial attacks on MNIST. ASR means attack success rate (\%). N.A. means ``not available'' due to zero ASR.}
 	\label{table_attack_mnist}
 \begin{tabular}{l|llll|llll|llll}
 	\hline
 	& \multicolumn{4}{c|}{Best case}    & \multicolumn{4}{c|}{Average case} & \multicolumn{4}{c}{Worst case}    \\ \hline
 	Attack           & ASR  & $L_1$ & $L_2$ & $L_\infty$ & ASR  & $L_1$ & $L_2$ & $L_\infty$ & ASR  & $L_1$ & $L_2$ & $L_\infty$ \\ \hline
 	C\&W ($L_2$)     & 100  & 13.93 & 1.377 & 0.379      & 100  & 22.46 & 1.972 & 0.514      & 99.9 & 32.3  & 2.639 & 0.663      \\
 	FGM-$L_1$        & 99.9 & 26.56 & 2.29  & 0.577      & 39   & 53.5  & 4.186 & 0.782      & 0    & N.A.  & N.A.  & N.A.       \\
 	FGM-$L_2$        & 99.4 & 25.84 & 2.245 & 0.574      & 34.6 & 39.15 & 3.284 & 0.747      & 0    & N.A.  & N.A.  & N.A.       \\
 	FGM-$L_\infty$   & 99.9 & 83.82 & 4.002 & 0.193      & 42.5 & 127.2 & 6.09  & 0.296      & 0    & N.A.  & N.A.  & N.A.       \\
 	I-FGM-$L_1$      & 100  & 18.34 & 1.605 & 0.403      & 100  & 32.94 & 2.606 & 0.591      & 100  & 53.02 & 3.979 & 0.807      \\
 	I-FGM-$L_2$      & 100  & 17.6  & 1.543 & 0.387      & 100  & 30.32 & 2.41  & 0.561      & 99.8 & 47.8  & 3.597 & 0.771      \\
 	I-FGM-$L_\infty$ & 100  & 49.8  & 2.45  & 0.147      & 100  & 71.39 & 3.472 & 0.227      & 99.9 & 95.48 & 4.604 & 0.348      \\
 	EAD (EN rule)    & 100  & 9.808 & 1.427 & 0.452      & 100  & 17.4  & 2.001 & 0.594      & 100  & 25.52 & 2.582 & 0.748      \\
 	EAD ($L_1$ rule) & 100  & 7.153 & 1.639 & 0.593      & 100  & 14.11 & 2.211 & 0.768      & 100  & 22.05 & 2.747 & 0.934      \\ \hline
 \end{tabular}
 \end{table*}

 \begin{table*}[]
 	\centering
 	\caption{Comparison of different adversarial attacks on CIFAR10. ASR means attack success rate (\%).}
 	\label{table_attack_cifar}
 \begin{tabular}{l|llll|llll|llll}
 	\hline
 	& \multicolumn{4}{c|}{Best case}    & \multicolumn{4}{c|}{Average case}  & \multicolumn{4}{c}{Worst case}    \\ \hline
 	Attack           & ASR  & $L_1$ & $L_2$ & $L_\infty$ & ASR  & $L_1$  & $L_2$ & $L_\infty$ & ASR  & $L_1$ & $L_2$ & $L_\infty$ \\ \hline
 	C\&W ($L_2$)     & 100  & 7.167 & 0.209 & 0.022      & 100  & 13.62  & 0.392 & 0.044      & 99.9 & 19.17 & 0.546 & 0.064      \\
 	FGM-$L_1$        & 99.5 & 14.76 & 0.434 & 0.049      & 48.8 & 51.97  & 1.48  & 0.152      & 0.7  & 157.5 & 4.345 & 0.415      \\
 	FGM-$L_2$        & 99.5 & 14.13 & 0.421 & 0.05       & 42.8 & 39.5   & 1.157 & 0.136      & 0.7  & 107.1 & 3.115 & 0.369      \\
 	FGM-$L_\infty$   & 100  & 32.74 & 0.595 & 0.011      & 52.3 & 127.81 & 2.373 & 0.047      & 0.6  & 246.4 & 4.554 & 0.086      \\
 	I-FGM-$L_1$      & 100  & 7.906 & 0.232 & 0.026      & 100  & 17.53  & 0.502 & 0.055      & 100  & 29.73 & 0.847 & 0.095      \\
 	I-FGM-$L_2$      & 100  & 7.587 & 0.223 & 0.025      & 100  & 17.12  & 0.489 & 0.054      & 100  & 28.94 & 0.823 & 0.092      \\
 	I-FGM-$L_\infty$ & 100  & 17.92 & 0.35  & 0.008      & 100  & 33.3   & 0.68  & 0.018      & 100  & 48.3  & 1.025 & 0.032      \\
 	EAD (EN rule)    & 100  & 4.014 & 0.261 & 0.047      & 100  & 8.18   & 0.502 & 0.097      & 100  & 12.11 & 0.69  & 0.147      \\
 	EAD ($L_1$ rule) & 100  & 2.597 & 0.359 & 0.103      & 100  & 6.066  & 0.613 & 0.17       & 100  & 8.986 & 0.871 & 0.27       \\ \hline
 \end{tabular}
 \end{table*}

 \begin{table*}[]
 	\centering
 	\caption{Comparison of different adversarial attacks on ImageNet. ASR means attack success rate (\%). N.A. means ``not available'' due to zero ASR.}
 	\label{table_attack_imagenet}
 	\begin{tabular}{l|llll|llll|llll}
 		\hline
 		& \multicolumn{4}{c|}{Best case}   & \multicolumn{4}{c|}{Average case} & \multicolumn{4}{c|}{Worst case}  \\ \hline
 		Method           & ASR & $L_1$ & $L_2$ & $L_\infty$ & ASR  & $L_1$ & $L_2$ & $L_\infty$ & ASR & $L_1$ & $L_2$ & $L_\infty$ \\ \hline
 		C\&W ($L_2$)     & 100 & 157.3 & 0.511 & 0.018      & 100  & 232.2 & 0.705 & 0.03       & 100 & 330.3 & 0.969 & 0.044      \\
 		FGM-$L_1$        & 9   & 193.3 & 0.661 & 0.025      & 1    & 61    & 0.187 & 0.007      & 0   & N.A.  & N.A.  & N.A.       \\
 		FGM-$L_2$        & 12  & 752.9 & 2.29  & 0.087      & 1    & 2338  & 6.823 & 0.25       & 0   & N.A.  & N.A.  & N.A.       \\
 		FGM-$L_\infty$   & 19  & 21640 & 45.76 & 0.115      & 3    & 3655  & 7.102 & 0.014      & 0   & N.A.  & N.A.  & N.A.       \\
 		I-FGM-$L_1$      & 98  & 292.2 & 0.89  & 0.03       & 77   & 526.4 & 1.609 & 0.054      & 34  & 695.5 & 2.104 & 0.078      \\
 		I-FGM-$L_2$      & 100 & 315.4 & 0.95  & 0.03       & 100  & 774.1 & 2.358 & 0.086      & 96  & 1326  & 4.064 & 0.153      \\
 		I-FGM-$L_\infty$ & 100 & 504.9 & 1.163 & 0.004      & 100  & 864.2 & 2.079 & 0.01       & 100 & 1408  & 3.465 & 0.019      \\
 		EAD (EN rule)    & 100 & 29.56 & 1.007 & 0.128      & 100  & 69.47 & 1.563 & 0.238      & 100 & 160.3 & 2.3   & 0.351      \\
 		EAD ($L_1$ rule) & 100 & 22.11 & 1.167 & 0.195      & 100  & 40.9  & 1.598 & 0.293      & 100 & 100   & 2.391 & 0.423      \\ \hline
 	\end{tabular}
 \end{table*}

 \subsection{Complete Results on Attacking Defensive Distillation on MNIST and CIFAR10 (Tables \ref{table_DD_mnist} and \ref{table_DD_cifar})}

 Tables \ref{table_DD_mnist} and \ref{table_DD_cifar} display the complete attack results of EAD and the C\&W method on breaking defensive distillation with different temperature parameter $T$ on MNIST and CIFAR10. Although defensive distillation is a standard defense technique for DNNs, EAD and the C\&W attack can successfully break defensive distillation with a wide range of temperature parameters.

 \begin{table*}[t]
 	\centering
 	\caption{Comparison of the C\&W method and EAD on attacking defensive distillation with different temperature parameter $T$ on MNIST. ASR means attack success rate (\%).}
 	\label{table_DD_mnist}
 \begin{tabular}{ll|llll|llll|llll}
 	\hline
 	&     & \multicolumn{4}{c|}{Best case}   & \multicolumn{4}{c|}{Average case} & \multicolumn{4}{c}{Worst case}    \\ \hline
 	Method                                                                       & $T$ & ASR & $L_1$ & $L_2$ & $L_\infty$ & ASR  & $L_1$ & $L_2$ & $L_\infty$ & ASR  & $L_1$ & $L_2$ & $L_\infty$ \\ \hline
 	\multirow{11}{*}{\begin{tabular}[c]{@{}l@{}}C\&W\\ ($L_2$)\end{tabular}}     & 1   & 100 & 14.02 & 1.31  & 0.347      & 100  & 23.27 & 1.938 & 0.507      & 99.8 & 32.96 & 2.576 & 0.683      \\
 	& 10  & 100 & 16.32 & 1.514 & 0.393      & 100  & 25.79 & 2.183 & 0.539      & 99.9 & 36.18 & 2.882 & 0.684      \\
 	& 20  & 100 & 16.08 & 1.407 & 0.336      & 99.9 & 26.31 & 2.111 & 0.489      & 99.6 & 37.59 & 2.85  & 0.65       \\
 	& 30  & 100 & 16.23 & 1.409 & 0.332      & 99.9 & 26.09 & 2.083 & 0.468      & 99.7 & 38.05 & 2.858 & 0.629      \\
 	& 40  & 100 & 16.16 & 1.425 & 0.355      & 100  & 27.03 & 2.164 & 0.501      & 100  & 39.06 & 2.955 & 0.667      \\
 	& 50  & 100 & 16.48 & 1.449 & 0.34       & 100  & 26.01 & 2.111 & 0.486      & 99.9 & 36.74 & 2.826 & 0.651      \\
 	& 60  & 100 & 16.94 & 1.506 & 0.36       & 100  & 27.44 & 2.247 & 0.512      & 99.7 & 38.74 & 2.998 & 0.668      \\
 	& 70  & 100 & 15.39 & 1.297 & 0.297      & 99.9 & 25.28 & 1.961 & 0.453      & 99.8 & 36.58 & 2.694 & 0.626      \\
 	& 80  & 100 & 15.86 & 1.315 & 0.291      & 100  & 26.89 & 2.062 & 0.46       & 99.9 & 38.83 & 2.857 & 0.651      \\
 	& 90  & 100 & 16.91 & 1.493 & 0.357      & 99.9 & 27.74 & 2.256 & 0.508      & 99.8 & 39.77 & 3.059 & 0.66       \\
 	& 100 & 100 & 16.99 & 1.525 & 0.365      & 99.9 & 27.95 & 2.3   & 0.518      & 99.8 & 40.11 & 3.11  & 0.67       \\ \hline
 	\multirow{11}{*}{\begin{tabular}[c]{@{}l@{}}EAD\\ (EN rule)\end{tabular}}    & 1   & 100 & 9.672 & 1.363 & 0.416      & 100  & 17.75 & 1.954 & 0.58       & 100  & 26.42 & 2.532 & 0.762      \\
 	& 10  & 100 & 11.79 & 1.566 & 0.468      & 100  & 20.5  & 2.224 & 0.617      & 100  & 29.59 & 2.866 & 0.768      \\
 	& 20  & 100 & 11.54 & 1.461 & 0.404      & 100  & 20.55 & 2.133 & 0.566      & 100  & 30.97 & 2.809 & 0.719      \\
 	& 30  & 100 & 11.82 & 1.463 & 0.398      & 100  & 20.98 & 2.134 & 0.546      & 100  & 31.92 & 2.847 & 0.701      \\
 	& 40  & 100 & 11.58 & 1.481 & 0.426      & 100  & 21.6  & 2.226 & 0.583      & 100  & 32.48 & 2.936 & 0.739      \\
 	& 50  & 100 & 12.11 & 1.503 & 0.408      & 100  & 21.09 & 2.161 & 0.56       & 99.9 & 30.52 & 2.806 & 0.731      \\
 	& 60  & 100 & 12.56 & 1.559 & 0.431      & 100  & 21.71 & 2.256 & 0.588      & 100  & 32.47 & 2.982 & 0.752      \\
 	& 70  & 100 & 11.1  & 1.353 & 0.356      & 100  & 19.92 & 2.001 & 0.518      & 99.9 & 29.99 & 2.655 & 0.691      \\
 	& 80  & 100 & 11.6  & 1.369 & 0.347      & 100  & 21.08 & 2.076 & 0.519      & 99.7 & 31.91 & 2.805 & 0.712      \\
 	& 90  & 100 & 12.57 & 1.546 & 0.424      & 100  & 22.62 & 2.312 & 0.587      & 99.9 & 33.38 & 3.047 & 0.744      \\
 	& 100 & 100 & 12.72 & 1.575 & 0.433      & 100  & 22.74 & 2.335 & 0.596      & 100  & 33.82 & 3.1   & 0.754      \\ \hline
 	\multirow{11}{*}{\begin{tabular}[c]{@{}l@{}}EAD\\ ($L_1$ rule)\end{tabular}} & 1   & 100 & 6.91  & 1.571 & 0.562      & 100  & 14    & 2.149 & 0.758      & 100  & 22.41 & 2.721 & 0.935      \\
 	& 10  & 100 & 8.472 & 1.806 & 0.628      & 100  & 16.16 & 2.428 & 0.798      & 99.9 & 25.38 & 3.061 & 0.945      \\
 	& 20  & 100 & 8.305 & 1.674 & 0.556      & 100  & 16.16 & 2.391 & 0.764      & 100  & 25.84 & 3.031 & 0.93       \\
 	& 30  & 100 & 8.978 & 1.613 & 0.51       & 100  & 16.48 & 2.417 & 0.749      & 100  & 25.72 & 3.124 & 0.932      \\
 	& 40  & 100 & 8.9   & 1.62  & 0.536      & 100  & 16.74 & 2.475 & 0.778      & 99.9 & 25.66 & 3.241 & 0.947      \\
 	& 50  & 100 & 9.319 & 1.645 & 0.51       & 100  & 17.01 & 2.401 & 0.744      & 100  & 24.59 & 3.103 & 0.939      \\
 	& 60  & 100 & 9.628 & 1.723 & 0.546      & 100  & 17.7  & 2.477 & 0.752      & 100  & 26.38 & 3.287 & 0.944      \\
 	& 70  & 100 & 8.419 & 1.524 & 0.466      & 100  & 16.01 & 2.193 & 0.674      & 100  & 24.31 & 2.935 & 0.892      \\
 	& 80  & 100 & 8.698 & 1.554 & 0.462      & 100  & 17.17 & 2.283 & 0.677      & 100  & 25.97 & 3.083 & 0.904      \\
 	& 90  & 100 & 9.219 & 1.755 & 0.557      & 100  & 18.46 & 2.529 & 0.744      & 99.9 & 27.9  & 3.315 & 0.926      \\
 	& 100 & 100 & 9.243 & 1.801 & 0.579      & 100  & 18.44 & 2.535 & 0.759      & 99.9 & 28.44 & 3.351 & 0.929      \\ \hline
 \end{tabular}
 \end{table*}

 \begin{table*}[t]
 	\centering
 	\caption{Comparison of the C\&W method and EAD on attacking defensive distillation with different temperature parameter $T$ on CIFAR10. ASR means attack success rate (\%).}
 	\label{table_DD_cifar}
 \begin{tabular}{ll|llll|llll|llll}
 \hline
                                                                               &     & \multicolumn{4}{c|}{Best case}   & \multicolumn{4}{c|}{Average case} & \multicolumn{4}{c}{Worst case}    \\ \hline
 Method                                                                        & $T$ & ASR & $L_1$ & $L_2$ & $L_\infty$ & ASR  & $L_1$ & $L_2$ & $L_\infty$ & ASR  & $L_1$ & $L_2$ & $L_\infty$ \\ \hline
 \multirow{11}{*}{\begin{tabular}[c]{@{}l@{}}C\&W \\ ($L_2$)\end{tabular}}     & 1   & 100 & 6.414 & 0.188 & 0.02       & 100  & 12.46 & 0.358 & 0.04       & 100  & 17.47 & 0.498 & 0.058      \\
                                                                               & 10  & 100 & 7.431 & 0.219 & 0.024      & 100  & 15.48 & 0.445 & 0.049      & 99.9 & 22.36 & 0.635 & 0.073      \\
                                                                               & 20  & 100 & 8.712 & 0.256 & 0.028      & 100  & 18.7  & 0.534 & 0.058      & 100  & 27.4  & 0.776 & 0.087      \\
                                                                               & 30  & 100 & 8.688 & 0.254 & 0.028      & 100  & 19.34 & 0.554 & 0.06       & 99.8 & 27.67 & 0.785 & 0.089      \\
                                                                               & 40  & 100 & 8.556 & 0.251 & 0.028      & 100  & 18.43 & 0.528 & 0.058      & 100  & 26.89 & 0.761 & 0.086      \\
                                                                               & 50  & 100 & 8.88  & 0.26  & 0.028      & 100  & 19.56 & 0.559 & 0.06       & 99.9 & 29.07 & 0.822 & 0.091      \\
                                                                               & 60  & 100 & 8.935 & 0.262 & 0.029      & 100  & 19.43 & 0.554 & 0.06       & 100  & 28.56 & 0.809 & 0.091      \\
                                                                               & 70  & 100 & 9.166 & 0.269 & 0.03       & 100  & 19.99 & 0.571 & 0.061      & 100  & 29.6  & 0.838 & 0.093      \\
                                                                               & 80  & 100 & 9.026 & 0.266 & 0.029      & 100  & 19.93 & 0.571 & 0.061      & 99.5 & 29.64 & 0.839 & 0.092      \\
                                                                               & 90  & 100 & 9.466 & 0.278 & 0.031      & 100  & 21.21 & 0.606 & 0.065      & 100  & 31.23 & 0.884 & 0.099      \\
                                                                               & 100 & 100 & 9.943 & 0.292 & 0.032      & 100  & 21.46 & 0.614 & 0.066      & 99.9 & 32.54 & 0.921 & 0.103      \\ \hline
 \multirow{11}{*}{\begin{tabular}[c]{@{}l@{}}EAD \\ (EN rule)\end{tabular}}    & 1   & 100 & 3.594 & 0.236 & 0.044      & 100  & 7.471 & 0.462 & 0.09       & 99.9 & 10.88 & 0.638 & 0.136      \\
                                                                               & 10  & 100 & 4.072 & 0.286 & 0.052      & 100  & 9.669 & 0.567 & 0.104      & 100  & 14.97 & 0.782 & 0.154      \\
                                                                               & 20  & 100 & 5.24  & 0.321 & 0.056      & 100  & 11.79 & 0.662 & 0.118      & 100  & 17.72 & 0.932 & 0.175      \\
                                                                               & 30  & 100 & 5.54  & 0.313 & 0.051      & 100  & 12.33 & 0.658 & 0.112      & 100  & 18.12 & 0.94  & 0.175      \\
                                                                               & 40  & 100 & 5.623 & 0.309 & 0.05       & 100  & 12.02 & 0.626 & 0.105      & 100  & 18.14 & 0.915 & 0.167      \\
                                                                               & 50  & 100 & 5.806 & 0.319 & 0.05       & 100  & 13    & 0.671 & 0.109      & 100  & 19.46 & 0.979 & 0.176      \\
                                                                               & 60  & 100 & 6.07  & 0.319 & 0.051      & 100  & 13.06 & 0.662 & 0.109      & 100  & 19.5  & 0.964 & 0.173      \\
                                                                               & 70  & 100 & 6.026 & 0.33  & 0.052      & 100  & 13.85 & 0.695 & 0.111      & 100  & 20.06 & 0.997 & 0.177      \\
                                                                               & 80  & 100 & 5.958 & 0.327 & 0.052      & 100  & 13.74 & 0.697 & 0.111      & 99.9 & 20.28 & 0.999 & 0.174      \\
                                                                               & 90  & 100 & 6.261 & 0.34  & 0.054      & 100  & 14.07 & 0.711 & 0.117      & 99.9 & 21.27 & 1.046 & 0.189      \\
                                                                               & 100 & 100 & 6.499 & 0.358 & 0.057      & 100  & 15.02 & 0.756 & 0.123      & 100  & 22.12 & 1.084 & 0.192      \\ \hline
 \multirow{11}{*}{\begin{tabular}[c]{@{}l@{}}EAD \\ ($L_1$ rule)\end{tabular}} & 1   & 100 & 2.302 & 0.328 & 0.098      & 100  & 5.595 & 0.556 & 0.154      & 99.9 & 8.41  & 0.776 & 0.237      \\
                                                                               & 10  & 100 & 2.884 & 0.37  & 0.101      & 100  & 6.857 & 0.718 & 0.2        & 100  & 10.44 & 1.037 & 0.321      \\
                                                                               & 20  & 100 & 3.445 & 0.435 & 0.118      & 100  & 8.943 & 0.802 & 0.201      & 99.9 & 13.73 & 1.164 & 0.322      \\
                                                                               & 30  & 100 & 3.372 & 0.452 & 0.128      & 100  & 8.802 & 0.796 & 0.202      & 100  & 14.31 & 1.124 & 0.304      \\
                                                                               & 40  & 100 & 3.234 & 0.465 & 0.136      & 100  & 8.537 & 0.809 & 0.209      & 100  & 13.82 & 1.093 & 0.297      \\
                                                                               & 50  & 100 & 3.402 & 0.479 & 0.142      & 100  & 8.965 & 0.848 & 0.22       & 100  & 14.96 & 1.17  & 0.317      \\
                                                                               & 60  & 100 & 3.319 & 0.497 & 0.151      & 100  & 8.647 & 0.863 & 0.232      & 99.8 & 14.46 & 1.174 & 0.325      \\
                                                                               & 70  & 100 & 3.438 & 0.506 & 0.153      & 100  & 9.344 & 0.913 & 0.239      & 100  & 15.06 & 1.214 & 0.331      \\
                                                                               & 80  & 100 & 3.418 & 0.5   & 0.15       & 100  & 9.202 & 0.914 & 0.242      & 100  & 15.14 & 1.227 & 0.337      \\
                                                                               & 90  & 100 & 3.603 & 0.519 & 0.157      & 100  & 9.654 & 0.96  & 0.258      & 99.8 & 15.82 & 1.287 & 0.361      \\
                                                                               & 100 & 100 & 3.702 & 0.543 & 0.161      & 100  & 9.839 & 0.993 & 0.269      & 99.8 & 16.22 & 1.351 & 0.379      \\ \hline
 \end{tabular}
 \end{table*}

 \subsection{Complete Attack Transferability Results on MNIST (Table \ref{table_transferability_full} )}
 Table \ref{table_transferability_full} summarizes the transfer attack results from an undefended DNN to a defensively distilled DNN on MNIST using EAD, the C\&W attack and I-FGM. I-FGM methods have poor performance in attack transferability. The average attack success rate (ASR) of  I-FGM is below 2\%. On the other hand, adjusting the transferability parameter $\kappa$ in EAD and the C\&W attack can significantly improve ASR. Tested on a wide range of $\kappa$ values,  the top average-case ASR for EAD is 98.6\% using the EN rule and 98.1\% using the $L_1$ rule. The top average-case ASR for the C\&W attack is 87.4\%. This improvement is significantly due to the improvement in the worst case, where the top worst-case ASR for EAD is 87\% using the EN rule and 85.8\% using the $L_1$ rule, while the top worst-case ASR for the C\&W attack is 30.5\%. The results suggest that $L_1$-based adversarial examples have better attack transferability.

 \begin{table*}[t]
 	\centering
 	\caption{Comparison of attack transferability from the undefended network to the defensively distilled network ($T=100$) on MNIST with varying transferability parameter $\kappa$. ASR means attack success rate (\%). N.A. means not ``not available'' due to zero ASR. There is no $\kappa$ parameter for I-FGM. }
 	\label{table_transferability_full}
 \begin{tabular}{ll|llll|llll|llll}
 	\hline
 	&          & \multicolumn{4}{c|}{Best case}    & \multicolumn{4}{c|}{Average case} & \multicolumn{4}{c|}{Worst case}    \\ \hline
 	Method                                                                        & $\kappa$ & ASR  & $L_1$ & $L_2$ & $L_\infty$ & ASR  & $L_1$ & $L_2$ & $L_\infty$ & ASR  & $L_1$  & $L_2$ & $L_\infty$ \\ \hline
 	I-FGM-$L_1$                                                                   & None     & 12.2 & 18.39 & 1.604 & 0.418      & 1.6  & 19    & 1.658 & 0.43       & 0    & N.A.   & N.A.  & N.A.       \\
 	I-FGM-$L_2$                                                                   & None     & 9.8  & 17.77 & 1.537 & 0.395      & 1.3  & 17.25 & 1.533 & 0.408      & 0    & N.A.   & N.A.  & N.A.       \\
 	I-FGM-$L_\infty$                                                              & None     & 14.7 & 46.38 & 2.311 & 0.145      & 1.7  & 48.3  & 2.44  & 0.158      & 0    & N.A.   & N.A.  & N.A.       \\ \hline
 	\multirow{13}{*}{\begin{tabular}[c]{@{}l@{}}C\&W\\ ($L_2$)\end{tabular}}      & 0        & 5.4  & 11.13 & 1.103 & 0.338      & 1.1  & 10.16 & 1.033 & 0.343      & 0    & N.A.   & N.A.  & N.A.       \\
 	& 5        & 16.6 & 15.58 & 1.491 & 0.424      & 3.4  & 17.35 & 1.615 & 0.46       & 0    & N.A.   & N.A.  & N.A.       \\
 	& 10       & 42.2 & 21.94 & 2.033 & 0.525      & 6.5  & 21.97 & 2.001 & 0.527      & 0    & N.A.   & N.A.  & N.A.       \\
 	& 15       & 74.2 & 27.65 & 2.491 & 0.603      & 22.6 & 32.54 & 2.869 & 0.671      & 0.4  & 56.93  & 4.628 & 0.843      \\
 	& 20       & 92.9 & 29.71 & 2.665 & 0.639      & 44.4 & 38.34 & 3.322 & 0.745      & 2.4  & 54.25  & 4.708 & 0.91       \\
 	& 25       & 98.7 & 30.12 & 2.719 & 0.664      & 62.9 & 45.41 & 3.837 & 0.805      & 10.9 & 71.22  & 5.946 & 0.972      \\
 	& 30       & 99.8 & 31.17 & 2.829 & 0.69       & 78.1 & 49.63 & 4.15  & 0.847      & 23   & 85.93  & 6.923 & 0.987      \\
 	& 35       & 100  & 33.27 & 3.012 & 0.727      & 84.2 & 55.56 & 4.583 & 0.886      & 30.5 & 105.9  & 8.072 & 0.993      \\
 	& 40       & 100  & 36.13 & 3.255 & 0.772      & 87.4 & 61.25 & 4.98  & 0.918      & 21   & 125.2  & 9.09  & 0.995      \\
 	& 45       & 100  & 39.86 & 3.553 & 0.818      & 85.2 & 67.82 & 5.43  & 0.936      & 7.4  & 146.9  & 10.21 & 0.996      \\
 	& 50       & 100  & 44.2  & 3.892 & 0.868      & 80.6 & 70.87 & 5.639 & 0.953      & 0.5  & 158.4  & 10.8  & 0.996      \\
 	& 55       & 100  & 49.37 & 4.284 & 0.907      & 73   & 76.77 & 6.034 & 0.969      & 0    & N.A.   & N.A.  & N.A.       \\
 	& 60       & 100  & 54.97 & 4.703 & 0.937      & 67.9 & 82.07 & 6.395 & 0.976      & 0    & N.A.   & N.A.  & N.A.       \\ \hline
 	\multirow{13}{*}{\begin{tabular}[c]{@{}l@{}}EAD \\ (EN rule)\end{tabular}}    & 0        & 6    & 8.373 & 1.197 & 0.426      & 0.6  & 4.876 & 0.813 & 0.307      & 0    & N.A.   & N.A.  & N.A.       \\
 	& 5        & 18.2 & 11.45 & 1.547 & 0.515      & 2.5  & 13.07 & 1.691 & 0.549      & 0    & N.A.   & N.A.  & N.A.       \\
 	& 10       & 39.5 & 15.36 & 1.916 & 0.59       & 8.4  & 16.45 & 1.989 & 0.6        & 0    & N.A.   & N.A.  & N.A.       \\
 	& 15       & 69.2 & 19.18 & 2.263 & 0.651      & 19.2 & 22.74 & 2.531 & 0.697      & 0.4  & 31.18  & 3.238 & 0.846      \\
 	& 20       & 89.5 & 21.98 & 2.519 & 0.692      & 37   & 28.36 & 2.99  & 0.778      & 1.8  & 39.91  & 3.951 & 0.897      \\
 	& 25       & 98.3 & 23.92 & 2.694 & 0.724      & 58   & 34.14 & 3.445 & 0.831      & 7.9  & 49.12  & 4.65  & 0.973      \\
 	& 30       & 99.9 & 25.52 & 2.838 & 0.748      & 76.3 & 40.2  & 3.909 & 0.884      & 23.7 & 59.9   & 5.404 & 0.993      \\
 	& 35       & 100  & 27.42 & 3.009 & 0.778      & 87.9 & 45.62 & 4.324 & 0.92       & 47.4 & 70.93  & 6.176 & 0.999      \\
 	& 40       & 100  & 30.23 & 3.248 & 0.814      & 95.2 & 52.33 & 4.805 & 0.945      & 71.3 & 83.19  & 6.981 & 1          \\
 	& 45       & 100  & 33.61 & 3.526 & 0.857      & 98   & 57.75 & 5.194 & 0.965      & 86.2 & 98.51  & 7.904 & 1          \\
 	& 50       & 100  & 37.59 & 3.843 & 0.899      & 98.6 & 66.22 & 5.758 & 0.978      & 87   & 115.7  & 8.851 & 1          \\
 	& 55       & 100  & 42.01 & 4.193 & 0.934      & 94.4 & 70.66 & 6.09  & 0.986      & 44.2 & 127    & 9.487 & 1          \\
 	& 60       & 100  & 46.7  & 4.562 & 0.961      & 90   & 75.59 & 6.419 & 0.992      & 13.3 & 140.35 & 10.3  & 1          \\ \hline
 	\multirow{13}{*}{\begin{tabular}[c]{@{}l@{}}EAD \\ ($L_1$ rule)\end{tabular}} & 0        & 6    & 6.392 & 1.431 & 0.628      & 0.5  & 6.57  & 1.565 & 0.678      & 0    & N.A.   & N.A.  & N.A.       \\
 	& 5        & 19   & 8.914 & 1.807 & 0.728      & 3.2  & 9.717 & 1.884 & 0.738      & 0    & N.A.   & N.A.  & N.A.       \\
 	& 10       & 40.6 & 12.16 & 2.154 & 0.773      & 7.5  & 13.74 & 2.27  & 0.8        & 0    & N.A.   & N.A.  & N.A.       \\
 	& 15       & 70.5 & 15.39 & 2.481 & 0.809      & 19   & 18.12 & 2.689 & 0.865      & 0.3  & 23.15  & 3.024 & 0.884      \\
 	& 20       & 90   & 17.73 & 2.718 & 0.83       & 39.4 & 24.15 & 3.182 & 0.902      & 1.9  & 38.22  & 4.173 & 0.979      \\
 	& 25       & 98.6 & 19.71 & 2.897 & 0.851      & 59.3 & 30.33 & 3.652 & 0.933      & 7.9  & 45.74  & 4.818 & 0.997      \\
 	& 30       & 99.8 & 21.1  & 3.023 & 0.862      & 76.9 & 37.38 & 4.191 & 0.954      & 22.2 & 55.54  & 5.529 & 1          \\
 	& 35       & 100  & 23    & 3.186 & 0.882      & 89.3 & 41.13 & 4.468 & 0.968      & 46.8 & 66.76  & 6.256 & 1          \\
 	& 40       & 100  & 25.86 & 3.406 & 0.904      & 96.3 & 47.54 & 4.913 & 0.979      & 69.9 & 80.05  & 7.064 & 1          \\
 	& 45       & 100  & 29.4  & 3.665 & 0.931      & 97.6 & 55.16 & 5.399 & 0.988      & 85.8 & 96.05  & 7.94  & 1          \\
 	& 50       & 100  & 33.71 & 3.957 & 0.95       & 98.1 & 62.01 & 5.856 & 0.992      & 85.7 & 113.6  & 8.845 & 1          \\
 	& 55       & 100  & 38.09 & 4.293 & 0.971      & 93.6 & 65.79 & 6.112 & 0.995      & 43.8 & 126.4  & 9.519 & 1          \\
 	& 60       & 100  & 42.7  & 4.66  & 0.985      & 89.6 & 72.49 & 6.572 & 0.997      & 13   & 141.3  & 10.36 & 1          \\ \hline
 \end{tabular}
 \end{table*}

 \subsection{Complete Results on Adversarial Training with $L_1$ and $L_2$ examples (Table \ref{table_adv_training_full})}

 Table \ref{table_adv_training_full} displays the complete results of adversarial training on MNIST using the $L_2$-based adversarial examples crafted by the C\&W attack and the $L_1$-based adversarial examples crafted by EAD with the EN or the $L_1$ decision rule. It can be observed that adversarial training with any single method can render the DNN more difficult to attack in terms of increased distortion metrics when compared with the null case. Notably, in the average case, joint adversarial training using $L_1$ and $L_2$ examples lead to increased $L_1$ and $L_2$ distortion against the C\&W attack and EAD (EN), and increased  $L_2$ distortion against EAD ($L_1$). The results suggest that EAD can complement adversarial training toward resilient DNNs. We would like to point out that in our experiments, adversarial training maintains comparable test accuracy. All the adversarially trained DNNs in Table \ref{table_adv_training_full} can still attain at least 99\% test accuracy on MNIST.

 \begin{table*}[t]
 	\centering
 	\caption{Comparison of adversarial training using the C\&W attack, EAD ($L_1$ rule) and  EAD (EN rule) on MNIST. ASR means attack success rate.}
 	\label{table_adv_training_full}
 \begin{tabular}{ll|llll|llll|llll}
 	\hline
 	&                                                                 & \multicolumn{4}{c|}{Best case}   & \multicolumn{4}{c|}{Average case} & \multicolumn{4}{c}{Worst case}    \\ \hline
 	\begin{tabular}[c]{@{}l@{}}Attack \\ method\end{tabular}                 & \begin{tabular}[c]{@{}l@{}}Adversarial \\ training\end{tabular} & ASR & $L_1$ & $L_2$ & $L_\infty$ & ASR  & $L_1$ & $L_2$ & $L_\infty$ & ASR  & $L_1$ & $L_2$ & $L_\infty$ \\ \hline
 	\multirow{4}{*}{\begin{tabular}[c]{@{}l@{}}C\&W \\ ($L_2$)\end{tabular}} & None                                                            & 100 & 13.93 & 1.377 & 0.379      & 100  & 22.46 & 1.972 & 0.514      & 99.9 & 32.3  & 2.639 & 0.663      \\
 	& EAD ($L_1$)                                                     & 100 & 15.98 & 1.704 & 0.492      & 100  & 26.11 & 2.468 & 0.643      & 99.7 & 36.37 & 3.229 & 0.794      \\
 	& C\&W ($L_2$)                                                    & 100 & 14.59 & 1.693 & 0.543      & 100  & 24.97 & 2.47  & 0.684      & 99.9 & 36.4  & 3.28  & 0.807      \\
 	& EAD + C\&W                                                      & 100 & 16.54 & 1.73  & 0.502      & 100  & 27.32 & 2.513 & 0.653      & 99.8 & 37.83 & 3.229 & 0.795      \\ \hline
 	\multirow{4}{*}{\begin{tabular}[c]{@{}l@{}}EAD \\ ($L_1$)\end{tabular}}  & None                                                            & 100 & 7.153 & 1.639 & 0.593      & 100  & 14.11 & 2.211 & 0.768      & 100  & 22.05 & 2.747 & 0.934      \\
 	& EAD ($L_1$)                                                     & 100 & 8.795 & 1.946 & 0.705      & 100  & 17.04 & 2.653 & 0.86       & 99.9 & 24.94 & 3.266 & 0.98       \\
 	& C\&W ($L_2$)                                                    & 100 & 7.657 & 1.912 & 0.743      & 100  & 15.49 & 2.628 & 0.892      & 100  & 24.16 & 3.3   & 0.986      \\
 	& EAD + C\&W                                                      & 100 & 8.936 & 1.975 & 0.711      & 100  & 16.83 & 2.66  & 0.87       & 99.9 & 25.55 & 3.288 & 0.979      \\ \hline
 	\multirow{4}{*}{\begin{tabular}[c]{@{}l@{}}EAD \\ (EN)\end{tabular}}     & None                                                            & 100 & 9.808 & 1.427 & 0.452      & 100  & 17.4  & 2.001 & 0.594      & 100  & 25.52 & 2.582 & 0.748      \\
 	& EAD (EN)                                                        & 100 & 11.13 & 1.778 & 0.606      & 100  & 19.52 & 2.476 & 0.751      & 99.9 & 28.15 & 3.182 & 0.892      \\
 	& C\&W ($L_2$)                                                    & 100 & 10.38 & 1.724 & 0.611      & 100  & 18.99 & 2.453 & 0.759      & 100  & 27.77 & 3.153 & 0.891      \\
 	& EAD + C\&W                                                      & 100 & 11.14 & 1.76  & 0.602      & 100  & 20.09 & 2.5   & 0.75       & 100  & 28.91 & 3.193 & 0.882      \\ \hline
 \end{tabular}
 \end{table*}

\end{document}